%% file: main.tex
\documentclass[conference,9pt]{IEEEtran}
\IEEEoverridecommandlockouts
\usepackage{booktabs} 
\usepackage{tabularx}
\usepackage{graphicx}
\graphicspath{{figure/}}
\usepackage{subcaption}
\usepackage{color}
\usepackage{comment}
\usepackage{xspace}
\usepackage{listings}
\usepackage{breqn}
\usepackage{balance}
\usepackage{siunitx}
\usepackage{diagbox}
\usepackage{enumitem} 
\newcommand{\reviews}[1]{#1}

\newcommand{\figautoref}[1]{{Fig.~\ref{#1}}}
\newcommand{\secautoref}[1]{{\S\ref{#1}}}

\makeatletter
  \def\sectionautorefname{\S\@gobble}
  \def\subsectionautorefname{\S\@gobble}
  \def\subsubsectionautorefname{\S\@gobble}

\makeatother

\usepackage[font={small,sf,bf}]{caption}

\newcommand{\mms}[1]{{\texttt{fairMS}}}
\newcommand{\dms}[1]{{\texttt{fairDS}}}
\newcommand{\proj}[1]{{\texttt{fairDMS}}}

\usepackage[noadjust]{cite}

\begin{document}

\title{fairDMS: Rapid Model Training by Data and Model Reuse}


\author{
\IEEEauthorblockN{
Ahsan Ali\IEEEauthorrefmark{1},
Hemant Sharma\IEEEauthorrefmark{2},
Rajkumar Kettimuthu\IEEEauthorrefmark{1},
Peter Kenesei\IEEEauthorrefmark{2},
Dennis Trujillo\IEEEauthorrefmark{2},\\
Antonino Miceli\IEEEauthorrefmark{2},
Ian Foster\IEEEauthorrefmark{1}\IEEEauthorrefmark{4}
Ryan Coffee\IEEEauthorrefmark{3},
Jana Thayer\IEEEauthorrefmark{3},
Zhengchun Liu\IEEEauthorrefmark{1}\IEEEauthorrefmark{4}
}
\IEEEauthorblockA{\IEEEauthorrefmark{1}Data Science and Learning Division,\
Argonne National Laboratory, Lemont, IL 60439, USA}
\thanks{Correspondence to: Zhengchun Liu, \texttt{<zhengchun.liu@anl.gov>}}
\IEEEauthorblockA{\IEEEauthorrefmark{2}X-ray Science Division, Argonne National Laboratory, Lemont, IL 60439, USA}
\IEEEauthorblockA{\IEEEauthorrefmark{3}SLAC National Accelerator Laboratory, Menlo Park, CA 94025, USA}
\IEEEauthorblockA{\IEEEauthorrefmark{4}University of Chicago, Chicago, IL 60637, USA}
}




\maketitle


\begin{abstract}
Extracting actionable information rapidly from data produced by instruments such as the Linac Coherent Light Source (LCLS-II) and Advanced Photon Source Upgrade (APS-U) is becoming ever more challenging due to high (up to TB/s) data rates.
Conventional physics-based information retrieval methods are hard-pressed to detect interesting events 
fast enough to enable timely focusing on a rare event or correction of an error. 
Machine learning~(ML) methods that learn cheap surrogate classifiers
present a promising alternative, but can fail catastrophically when changes in instrument or sample result in degradation in ML performance. 
To overcome such difficulties, we present a new data storage and ML model training architecture designed to organize large volumes of data and models
so that when model degradation is detected, prior models and/or data can be queried rapidly and a more suitable model retrieved and fine-tuned for new conditions. 
We show that our approach can achieve up to 100x data labelling speedup compared to the current state-of-the-art, 200x improvement in training speed, and 92x speedup in-terms of end-to-end model updating time. 
\end{abstract}

\maketitle
\pagestyle{plain}

\input{motivation.tex}
\input{methodology.tex}
\input{results.tex}
\input{discussion}
\input{related.tex}
\section{Conclusion}\label{sec:conc}

We have presented a step towards a fully automated ML pipeline for actionable information retrieval from data generated at high volume and velocity data sources. 
Our proposed solution, \proj{}, is an end-to-end ML framework designed for management of high velocity scientific data and rapid ML model training. 
It comprises two primary components, a data management service platform, \dms{}, and a model management service platform, \mms{}, that together allow it to adapt ML models to new data at greatly reduced cost relative to conventional methods.
It does this by reducing both data labeling and model retraining times. 
\dms{} leverages specialized data indexing methods to reduce data annotation time by reusing annotations generated previously for similar data; in so doing, it overcomes a major bottleneck in ML model training workflows. 
\mms{} uses the representation-building functionality of \dms{} to identify previously trained ML models that can serve as a good foundation for fine-tuning; in so doing, it allows retraining to proceed far more efficiently than otherwise.
Our experiments on two representative ML-based scientific applications show that \proj{} provides more than 50x speedup in the worst case and nearly 600x in the best case on ML model updating. 

\section*{Acknowledgment}
This material was based upon work supported by the U.S. Department of Energy, Office of Science, under contract DE-AC02-06CH11357 and Office of Basic Energy Sciences under Award Number FWP-35896.
This research used resources of the Argonne Leadership Computing Facility, a DOE Office of Science User Facility supported under Contract DE-AC02-06CH11357.
We thank the three anonymous referees for their detailed and constructive comments and suggestions, which have helped us clarifying lots of confusing statements. 
We also want to thank our shepherd, Thomas Peterka, for helping us to revise the paper.

\balance

\bibliographystyle{abbrv}
\bibliography{refs}

\section*{Government License}
The submitted manuscript has been created by UChicago Argonne, LLC, Operator of Argonne National Laboratory (``Argonne''). Argonne, a U.S.\ Department of Energy Office of Science laboratory, is operated under Contract No.\ DE-AC02-06CH11357. The U.S.\ Government retains for itself, and others acting on its behalf, a paid-up nonexclusive, irrevocable worldwide license in said article to reproduce, prepare derivative works, distribute copies to the public, and perform publicly and display publicly, by or on behalf of the Government.  The Department of Energy will provide public access to these results of federally sponsored research in accordance with the DOE Public Access Plan. http://energy.gov/downloads/doe-public-access-plan.

\end{document}

%% file: motivation.tex
\section{Introduction}\label{methodology}

Data generated by experiments, simulations, and digital twins, and by machine learning (ML) models derived from those data, 
are used on multiple time and distance scales. 
For example, data from an in-situ experiment may need to be delivered quasi-instantaneously to an ML model trainer so that it can rapidly update the digital twin that is to be used to choose the next experiment~\cite{liu2021bridge}. 
In this setting, it is vital that we be able to identify and deliver quickly both the best historical model to use as a basis for the digital twin and, for fine tuning of that model, the data most relevant to a specific training scenario. 
Data and trained models can also have value to other scientists, e.g., when designing and steering subsequent experiments and to construct and update other ML models, and thus must also be efficiently accessible to those other parties. 
In other words, we want to render  large quantities of scientific data, and also ML models trained on those data, findable, accessible, interoperable, and reusable (FAIR)~\cite{wilkinson2016fair}. 
To this end, we propose a new \textbf{FAIR Data and Model Service} (\proj{}) to provide \textit{indexing}, \textit{publication}, \textit{enrichment}, \textit{discovery}, and \textit{access} of both data and trained ML models used within ML-based scientific applications.

\begin{figure}[htb]
\includegraphics[width=\columnwidth]{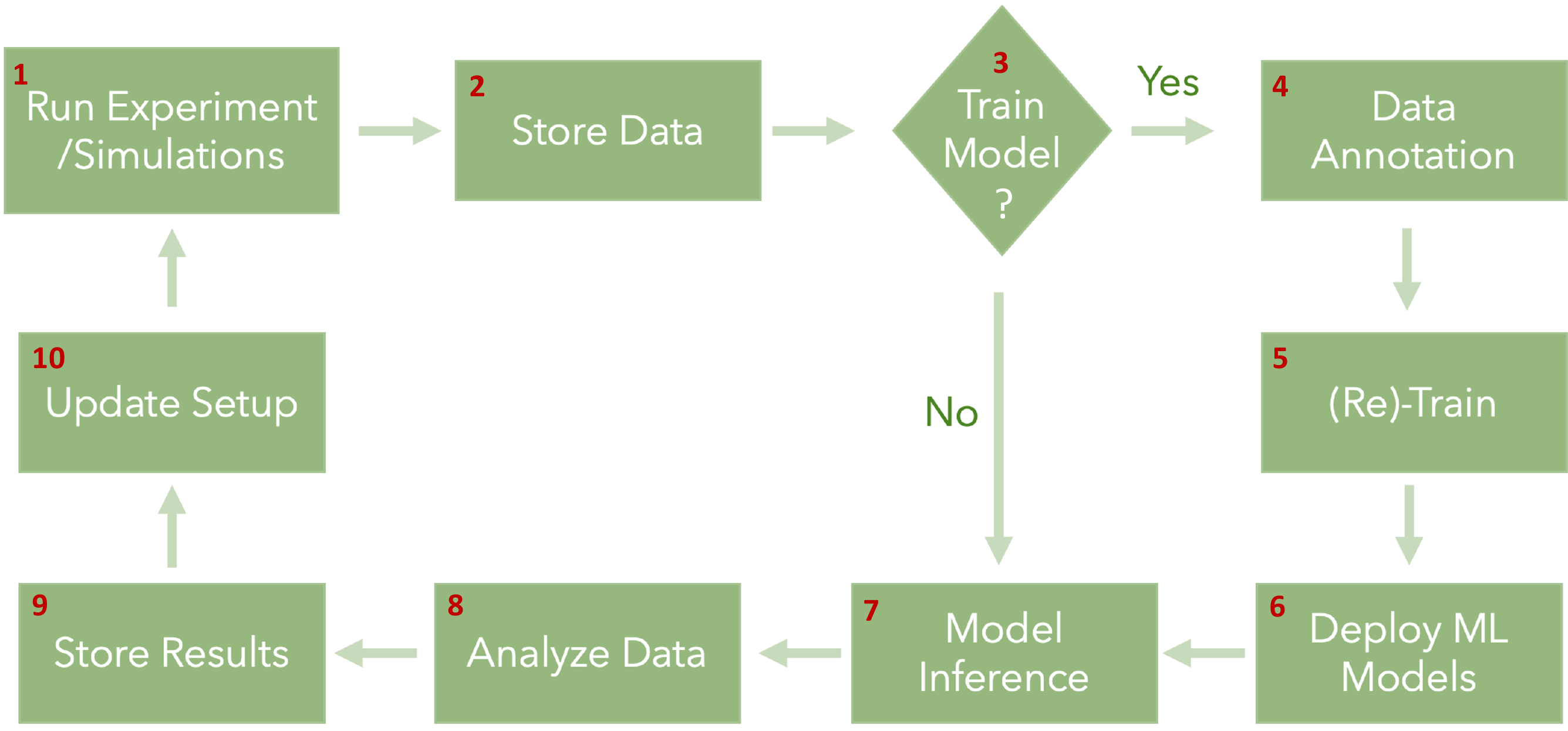}
\caption{An end-to-end scientific application workflow.}
\label{fig:sa_workflow}
\end{figure}

Experiments conducted at the Advanced Photon Source (APS) and Linac Coherent Light Source (LCLS) to study rare events, such as crack initiation and phase transformations, or weak processes like nonlinear X-ray methods, can generate data at up to TB/s \cite{Audrey2019}. 
An appropriately trained ML model can often extract the actionable information from such data streams significantly faster than traditional analytical methods~\cite{liu2021bridge,liu2020tomogan,BraggNN-IUCrJ,liu2019deep}. 
We show in \figautoref{fig:sa_workflow} a representative example of an ML-embedded scientific experiment workflow.
In (1), a scientific experiment or simulation is run with a particular setting, generating data that are (2) stored for further analysis.
Depending on a decision process (3) that may assess 
To extract actionable information at a rapid rate by using AI, the stored data is further (4) annotated by using conventional methods such as Pseudo-voigt to compute labels \cite{sharma2012fast}. 
A ML model is (5) trained on the annotated data, and the newly trained model is (6) deployed for (7) inference to obtain valuable information, such as peak location and shape in the case of High Energy X-ray Diffraction Microscopy (HEDM), from subsequent data. 
These results are (8, 9) stored and analyzed by the domain scientist. 
Based on the information gathered during this step, the domain scientist (10) updates the experimental setup or simulation parameters and the cycle continues. 

Changes in experiment or simulation parameters can cause variation in the characteristics of the data generated in (1) that result in declining ML model performance in (7). 
In the system demonstrated in \figautoref{fig:sa_workflow}, this problem of deteriorating ML model performance is addressed by repeating steps (4) and (5) after every experiment.
However, as those latter steps can each take hundreds of seconds, this approach leads to inefficient use of expensive scientific apparatus, and hinders tracking of dynamic behaviors.  

\begin{figure}[htb]
\centering
\includegraphics[width=\columnwidth]{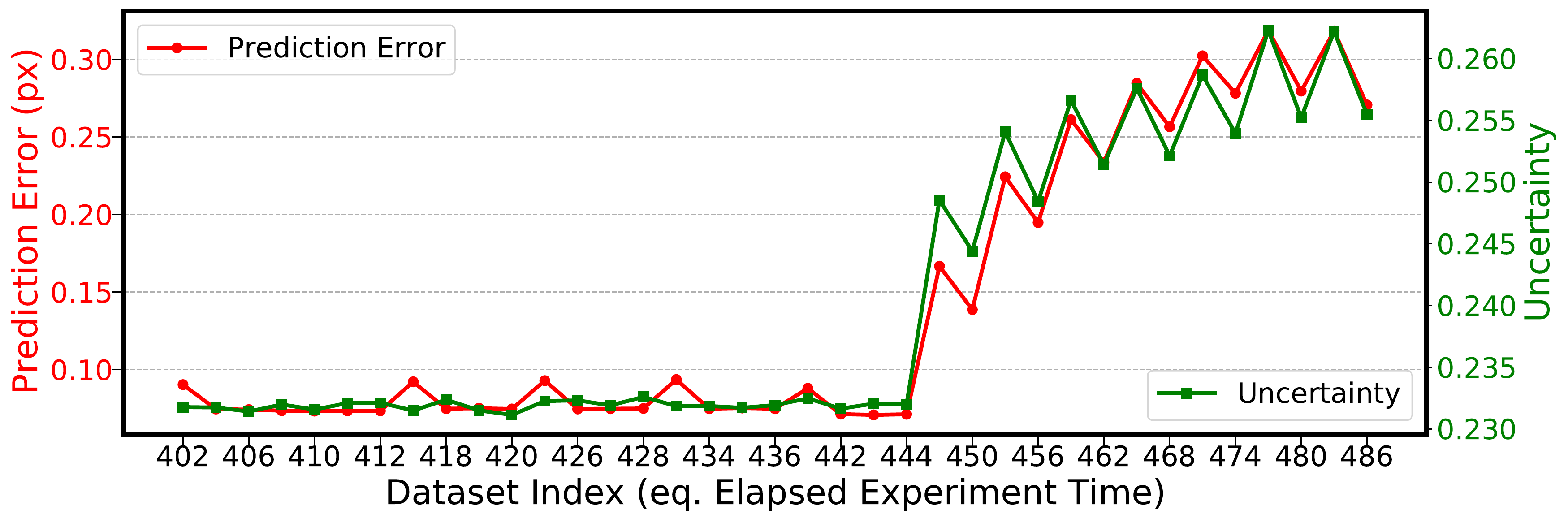}
\caption{Prediction error and uncertainty quantification for a ML model trained with data generated in the early stages of a HEDM experiment. 
X-axis: Sequence of data acquisition (equivalent to elapsed experiment time). Left y-axis: Prediction error. Right y-axis: 95\% confidence bound, quantified by using MC Dropout~\cite{gal2016dropout}.}
\label{fig:braggnn-uq}
\end{figure}

To illustrate model degradation over time, we show in \figautoref{fig:braggnn-uq}, in red, prediction error for an ML model trained on data generated in the first phase of a HEDM experiment, i.e., up to scan 402. 
The resulting model performs effectively until scan 444, after which its performance starts to deteriorate due to sample deformation. 
While such deformations are familiar phenomena in HEDM experiments, the researcher typically does not know, in realtime, when they will occur and thus cannot plan ahead.

Thus we want methods that will allow for rapid updates to ML models at the time of model degradation.
The rapid model update problem is in fact three distinct problems:
a) determining when model performance degrades, and thus model updating is needed;
b) generating labels for the new data produced in step (1) of \figautoref{fig:sa_workflow}, and 
c) retraining/updating the model.
The last task, model retraining, can be accelerated significantly by the use of purpose-built AI systems~\cite{liu2021bridge}, but
the first two tasks are more difficult.
\textit{We propose in this paper new methods that address these challenges by first, detecting when model performance degrades; second, identifying data from previous experiments that can be reused to reduce labeling efforts} (pseudo-labeling, handled by our FAIR data service, \textbf{\dms{}}); \emph{and third, for finding a suitable model, from a Zoo of models trained for previous experiments, as a foundation to fine-tune with new data to improve its performance}: our FAIR model service, \textbf{\mms{}}.


\subsection{The FairDS Data Service}
Scientific experiments can generate large quantities of data, but those data are commonly not well labeled. 
Existing scientific image repositories (e.g., TomoBank~\cite{de2018tomobank}, digital rocks portal~\cite{digital_rocks_portal}, PSI public data repository~\cite{spurin2020real}, the Materials Data Facility~\cite{blaiszik2016materials}) typically rely on human-supplied annotations to enable navigation via exogenous (e.g., namespace- or metadata-based) queries~\cite{de2018tomobank,digital_rocks_portal,spurin2020real}.
However, although human-supplied annotation makes data indexing and querying easy, it dos not scale, discourages data contributions, 
and can lead to biased labels.

A key novelty of our proposed \textbf{FAIR data service}, \dms{}, is its focus on \textit{fully automated feature extraction} via the use of (self-)supervised and unsupervised approaches to transform bulky, redundant image representations into compact, semantic-rich representations of visually salient characteristics.
As we will show, this transformation permits subsequent \textit{rapid data discovery and retrieval} (e.g., of images) based on proximity within compact feature spaces---a capability that we argue is essential for today's data-intensive and ML-based workflows, but is under-served by traditional data management systems.


\subsection{The FairMS Model service}
When one is given a ML task, such as using a new labeled dataset of X-ray diffraction images to construct a predictor of Bragg peak locations, we might reasonably decide to use the new dataset to train a new network (e.g., a deep neural network, DNN) \textit{from scratch}~\cite{BraggNN-IUCrJ}. 
However, the millions of parameters in typical DNNs can easily result in overfitting---particularly when the new training dataset is small. 
An alternative approach can then be considered.
In many settings, the new dataset is not dramatically different from datasets previously generated for the same or similar experiments, and thus models previously trained on those previous datasets have already learned features relevant to the new dataset.
Thus, we can apply a technique called \textit{fine tuning}, in which we unfreeze all or part of a previous model and re-train it on the new data (e.g., using a much smaller learning rate).
More specifically, we start with a network trained on a large historic dataset that we then fine-tune repeatedly as a succession of smaller datasets are acquired during an experiment. 
This approach can potentially achieve meaningful improvements, by incrementally adapting the pre-trained features to the new data.


\reviews{Researchers have previously developed data management solutions for ML training workloads~\cite{ wang2020diesel,ren2012tablefs,ren2014indexfs}, methods for labeling training data~\cite{bai2021self} and quantifying similarity between data points with self-supervised learning methods~\cite{grill2020bootstrap,bengio2014deep,wang2015unsupervised}, and model recommendation systems~\cite{li2018ease}.
However, to the best of our knowledge, no existing framework addresses the problems of training data management, automatic data lookup, and model recommendation under a single platform.} 
In this work we aim to address the challenges related to training ML models for in-situ actionable information retrieval for scientific applications that generate high volume and velocity data. 

\subsection{FairDS + FairMS = FairDMS}

To this end, we propose an end-to-end framework, \proj{}, that is composed of a \dms{} component that uses representation learning to generate robust and efficient data labels and a \mms{} component that indexes trained ML models, by a learnt representation of the training dataset, to recommend the best model as a foundation for efficient fine-tuning (i.e., fine tuning with rapid convergence) for ML model training. 
Our main contributions are as follows:
\begin{itemize}
\item A data management system, \dms{}, that leverages historical data through  self-supervised learning for pseudo-labeling high velocity scientific data to reduce the data engineering efforts, and human agent feedback during the scientific experiments.
\item An autonomous model indexing and recommendation system that takes the user data as an input and recommends an ML model, from a Zoo of models trained in the past for datasets of the similar experiment, as a foundation for fine-tuning to reduce the model training time. 
\item We further implemented our rapid ML model training system and evaluated the performance using three representative scientific datasets and two different scientific applications.
\end{itemize}

The remainder of the paper is organized as follows: In \secautoref{sec:introduction} we introduce the \proj{} architecture and describe its modules and their functionality. We further evaluate \proj{} performance in \secautoref{results}, and \reviews{in \secautoref{limits} we discuss its limitations}. Finally, we review the current state of the art in \secautoref{related} and conclude in \secautoref{sec:conc}. 

%% file: methodology.tex
\section{Key insights and contributions} \label{sec:introduction}
\begin{figure}[htb]
\centering
\includegraphics[width=\linewidth]{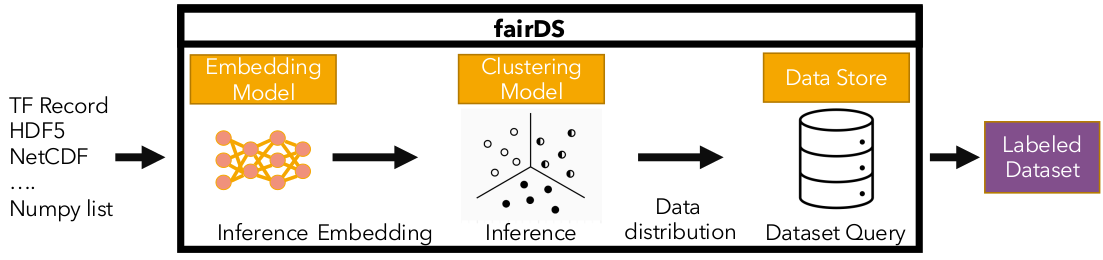}
\caption{Architecture and building blocks of the data management and service platform.}
\label{fig:overview_data_repo}
\end{figure}

We summarize in Figs.~\ref{fig:overview_data_repo} and~\ref{fig:overview_model_repo} the internal architectures of the \dms{} and \mms{} building blocks, respectively, that we use to construct our fairDMS, and show in \figautoref{fig:dms-for-workflow} the end-to-end architecture of the \proj{} system.
In the following, we describe the key functions and features of each module and discuss how each contributes towards the successful implementation of \proj{}.

\subsection{Data Management and Service Platform} \label{data_repo}

We have noted that a major challenge when applying an ML model to data from a high-rate scientific experiment is that as experimental conditions change, new data can become dissimilar to the data used for model training, in which model predictions become suspect.
The obvious solution is to retrain the ML model on new labeled data. And indeed we have new data, from recent experiments.
However, we do not have labels and the only way that we have of generating accurate labels for new data, namely running physics-based computational procedures, is expensive.

The fairDS component of fairDMS aims to enable an alternative solution to this problem of obtaining labeled data in which instead of labeling new data we find similar, already labeled data from past experiments. 
%
To this end, we apply a form of content-based data retrieval~\cite{kokare2002survey},
an approach to searching large image collections based on 
similarity to sample images. 

One approach to retrieving labeled historical data that are similar to unlabeled input data would be to perform pixel-by-pixel intensity vector comparisons of image pairs with a distance metric such as Nearest Neighbour---also referred to as instance discrimination \cite{chen2020simple}. 
However, this approach is both fragile (e.g., it is sensitive to the location of individual pixels and cannot easily identify similar but translated, scaled, or rotated images) and expensive (its computational cost scales linearly with the size of the database).
To overcome these challenges, we instead use self-supervised learning methods to generate compact representations (\textit{embeddings}) of individual images and then unsupervised clustering methods to group samples with similar representations into a specified number of clusters: see \figautoref{fig:overview_data_repo}.
Once these methods have been applied to historical data, it is straightforward to take new, unlabeled data as input, compute its embedding, and retrieve 
a group of similar historical data based upon the embedding.


ML algorithms for self-supervised learning of compact embeddings are used widely for extracting key features of input data that represent discrete variables as continuous vectors~\cite{jansen2017word,guo2016entity}. 
The key idea is that not all pixels in an image provide important information hence, only the most relevant features should be extracted for comparison to find the similar data. 
Self-supervised representation learning is generally formulated as learning an embedding (i.e., a feature vector for each sample) such that images that are semantically similar are close in embedding space, while semantically different ones are far apart. 
The goal is to generate an embedding of the images that solve the ``pretext" task and are also generalizable for other unseen samples. 
Various representation learning algorithms are discussed in \secautoref{related}. 
Depending on specific application and data, different algorithms can be used for \dms{}.

The low-dimensional feature representation generated by the embedding module enables to capture the most important features of the input image without the involvement of a human agent.
In addition, the smaller size of the embedding representation relative to the original experimental data reduces computational costs during the lookup operation. 
A third advantage is that it allows \dms{} to find similar labeled images even when subject to various transformations, such as shifting, rotations, and mirroring. 



Having computing embeddings, the \textit{Clustering module} takes the low dimension representation of each sample as an input and further assigns each input into a cluster by using a clustering algorithm so as to enable two-level hierarchical search (i.e., first find the cluster, then find the most similar sample within the cluster).  
We use K-means clustering~\cite{duda1973pattern} for the experiments performed in this study due to its scalability and fast convergence compared to other clustering algorithms. 
The k-means clustering algorithm stores the input samples as  points in the feature space and further calculates the distance using normalized Euclidean distance between $K$ cluster centers so as to assign each sample to a nearest cluster center. 
A challenge when using K-means clustering is selecting an appropriate number of clusters $K$.
We employ the \textit{elbow method}~\cite{yuan2019research} to select $K$ for a specific applications and dataset automatically. 
This method is based on the observation that for a particular application and dataset, as the number of clusters increases the Within-cluster-Sum of Squared errors (WSS) inevitably decreases, but at a rate that diminishes for the optimal number of clusters. 
We use the YellowBrick~\cite{Yellowbrick} library for optimizing the value of $K$ in an automatic fashion. 

\textit{Data Store:} Once the input dataset has been transformed by the \textit{Embedding Module} and assigned to clusters by the \textit{Clustering Module}, \dms{} generates a cluster probability distribution function (PDF) for the input dataset, i.e, the probability that images belong to a certain cluster.  
Then, \dms{} queries the data store, which contains the labeled historical data, along with embedding information of each sample and its respective cluster ID. 
The data store takes the PDF of the input data and generates from the historical data, a labeled dataset with similar characteristics to the input data. 
It returns the same number of labeled images as are in the input data, selected randomly from each cluster based on the PDF of the input dataset. 

Key requirements for the \textit{Data Store} module are that it:
i) scale to store large amount of data; 
ii) provide efficient data look up by using embedding indexing, in order to minimize the labeling time; 
iii) support data updates for adding newly labeled data;
iv) support parallel reads during the training phase; and 
v) allow parallel writes during the data update phase. 
Keeping these characteristics in mind, we adopt MongoDB~\cite{MongoDB} as the \dms{} data store solution.
MongoDB, an open source NoSQL database that offers high availability and scalability,
supports both vertical and horizontal scaling.
It stores unstructured or semi-structured data in JSON-like documents with optional schema. 
It also supports efficient data updates and reading via indexing~\cite{MongoIndexing}.

In summary, \dms{} first employs an efficient data indexing mechanism to generate an optimized storage layout for historical labeled data. 
Next, it provides high degrees of parallelism for fast data access to minimize the I/O overhead during model training. 
It further minimize labeling time by building data indexes as data are written to storage. 
\dms{} enables robust labeled data lookup for a given unlabeled datum, from the historical data. 

\subsection{Model Management and Service Platform}\label{model_repo}
Over time, \mms{} accumulates many instances of the same model architecture, each trained on different training data: what we refer to below as a Zoo of models. 
Any of these models can be fine tuned with new data to obtain a fine-tuned model.
We discuss here how \mms{} enables model recommendations for fine-tuning that reduce both training time and resource consumption. 
The key idea is to \textit{use the learned distribution of the training dataset to index a trained ML model}.

Fine tuning \cite{guyon2011unsupervised,donahue2014decaf} and transfer learning \cite{pan2009survey,donahue2014decaf} are widely used in ML.    
The key idea is that when faced with the task of determining parameters $P$ for a model $M$ that generate accurate predictions for a new dataset $D$, it can be far more efficient to use a fine-tuning process to adjust the parameters $P'$ obtained when  $M$ was previously trained on another dataset $D'$, rather than training $M$ from scratch on $D$.
If $D'$ (and thus $P'$) are appropriately chosen, the fine-tuning process can converge to good parameters $P$ far faster, and with the use of many fewer resources, than if $M$ was trained from scratch.

A challenge when seeking to apply these methods in scientific applications is identifying the proper model to be used as the foundation. 
Simply choosing a model at random, even when trained with data from the same experiment, may not result in any improvement in training time (see discussion in \figautoref{fig:e2e_CookieNetAE} and \figautoref{fig:e2e_BraggNN}). 
Hence, it is necessary to create meaningful representations for each model in the Zoo so as to index them properly. 

\begin{figure}[htb]
\centering
\includegraphics[width=\linewidth,trim=12mm 2mm 2mm 5mm,clip]{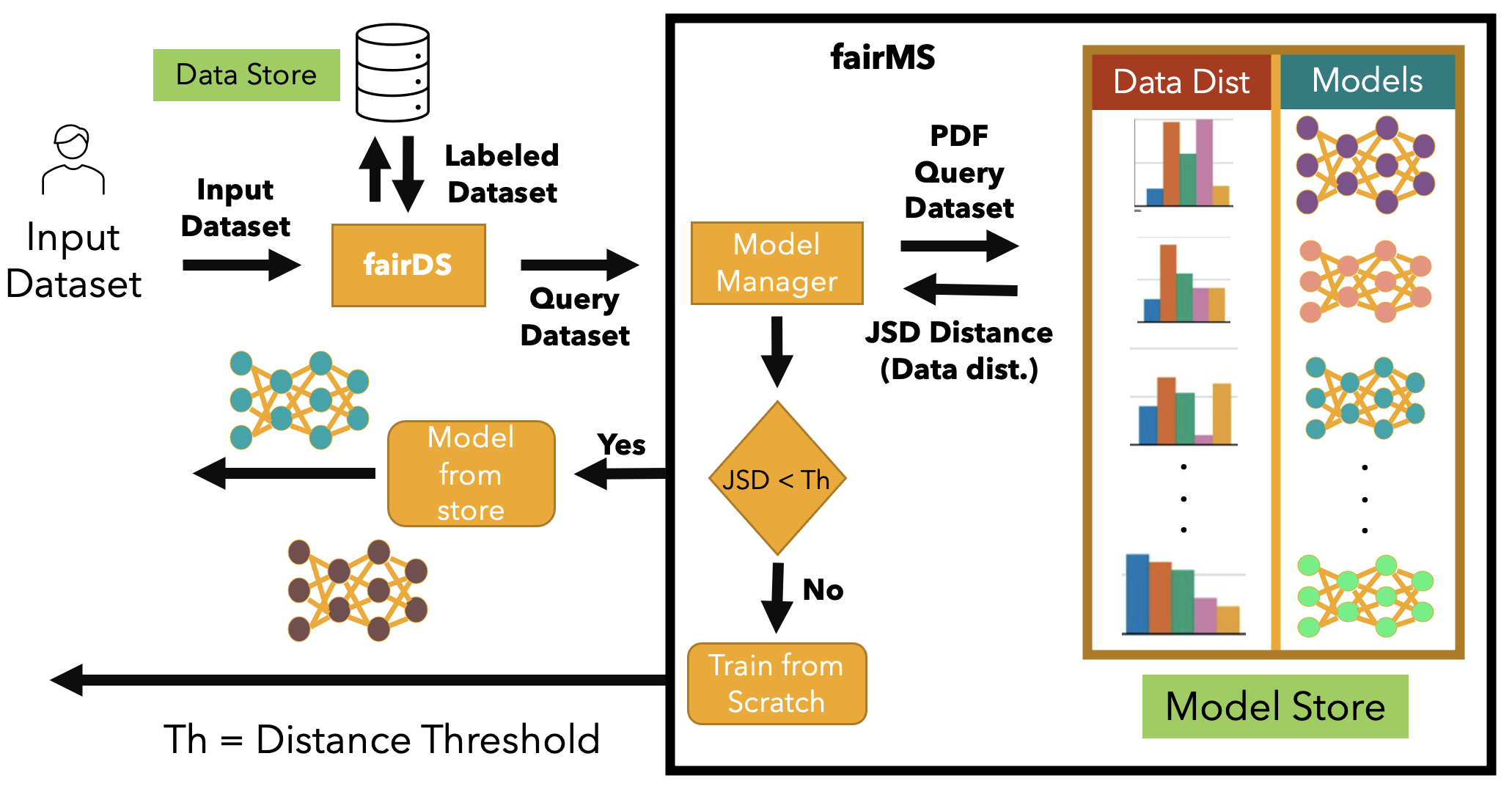}
\caption{Architecture and building blocks of the model management and service platform.}
\label{fig:overview_model_repo}
\end{figure}

\figautoref{fig:overview_model_repo} illustrates the model management service and its associated modules.
With an input dataset as key, \mms{} first uses \dms{} to determine its distribution representation as a PDF.
It then compares this distribution with that of each training dataset in the model Zoo, via Jensen–Shannon divergence~\cite{fuglede2004jensen}, so as to identify the  
model that was trained with the training dataset closest to the input dataset.
This model is return as the foundation model to be used for fine-tuning. 
We evaluate the effectiveness of this approach to model retrieval for fine tuning in \secautoref{sec:MSval}. 

Next, we discuss the functionality of the Model Zoo and Model manager components of the \mms{}.
The \textit{Model Zoo} is responsible for managing all models that have previously been trained with the various datasets generated in past experiments. 
In order to enable selection of appropriate models for fine-tuning, Model Zoo tracks for each such model its training data distribution (based on our clustering indexing: see \secautoref{data_repo}), as shown in \figautoref{fig:overview_model_repo}. 
This data distribution information allows \mms{} to find the best model for fine-tuning without a need to run any model inference. 

The \textit{Model Manager} takes the distribution (as generated by \dms{}) for a new user-supplied dataset as input and calculates the similarity, using \textit{Jensen–Shannon divergence} (JSD) \cite{fuglede2004jensen}, between this data distribution and the distribution recorded for each model in the Zoo. 
The JSD, a principled divergence measure between two probability distributions (also known as information radius), quantifies the similarity among two or more distributions. 
Its value is bounded by 0 and 1 for two probability distributions, with 0 indicating completely similar distributions and 1 indicating orthogonal distributions.




\subsection{Rapid Model Training Workflow}
The gradient descent method used in deep learning training is an iterative optimization algorithm that is commonly performed to \textit{convergence}: that is, until such time as model error (as evaluated on test data) no longer declines.  
Training can start from any training checkpoint: for example, from a partially trained model (i.e., a model not trained to convergence) or from a model trained to convergence on similar data. 
When fine tuning a previous trained model on new data, a key figure of merit is how much time and what computational resources are required for the fine tuning process to converge. 

Model convergence time can be impacted by the co-relation between the new data and the model checkpoint. 
If the checkpoint model was trained on data with similar characteristics to the new data, the training can be expected to converge in fewer iterations than if its training data were dissimilar~\cite{guyon2011unsupervised,donahue2014decaf}. 
Hence, we use \mms{} to identify the model in the Zoo with a training dataset that is most similar, by our JSD metric, to the new data that is to be used for fine tuning.
We also apply a user-defined distance threshold: if no historical model dataset is within that distance of the new dataset, a model will be trained from scratch.

If a new model is trained, it is both transferred to the user and  added to the model Zoo along with its training dataset distribution. 
Hence, the model Zoo can respond with this model in the future if presented with a similar data distribution. 

\begin{figure}[htb]
\centering
\includegraphics[width=\linewidth]{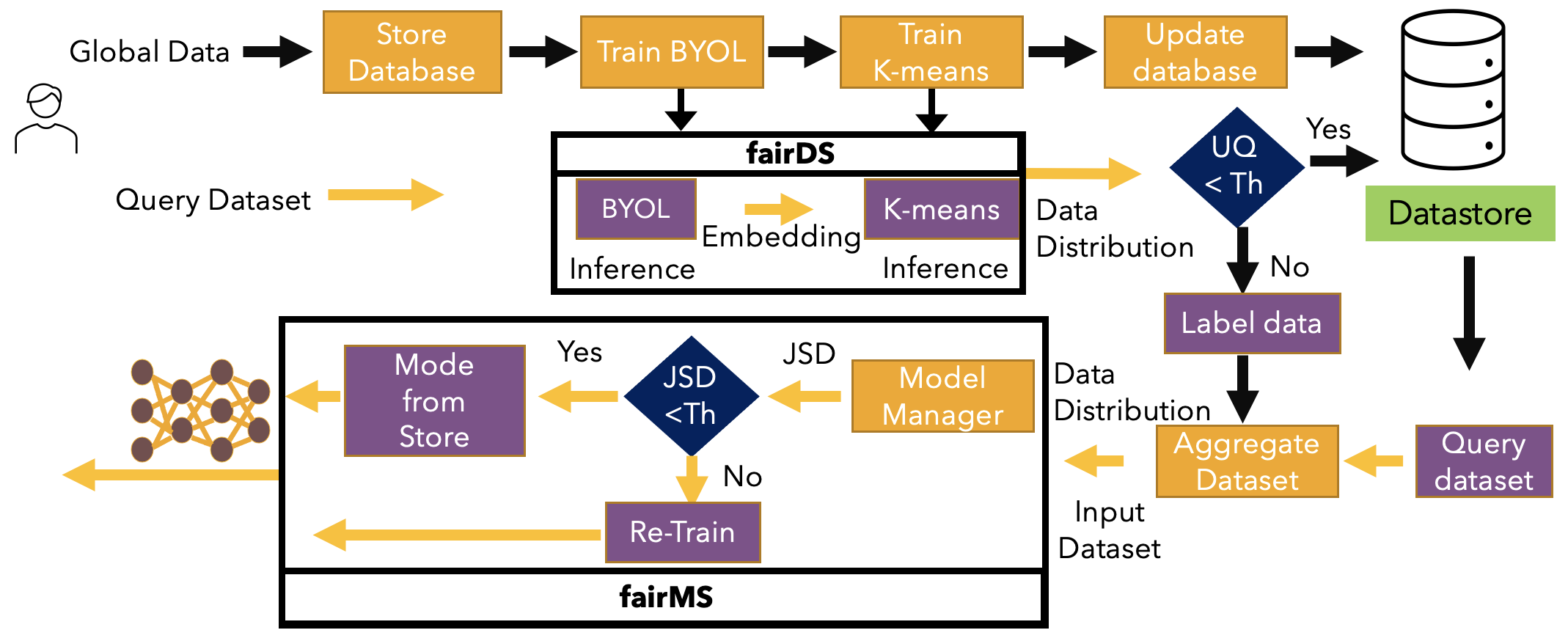}
\caption{\proj{} architecture and building blocks based on \dms{} and \mms{}.}
\label{fig:dms-for-workflow}
\end{figure}

\figautoref{fig:dms-for-workflow} shows how we combine the \texttt{fairDS} and \texttt{fairMS} components to construct our
rapid model training system. 
We can usefully distinguish, when describing this system, between 
\textit{user plane} operations that an end user can execute (or has direct access to)---colored purple in \figautoref{fig:dms-for-workflow}---and \textit{system plane} operations that are performed in the background without the involvement of the end user: colored yellow in the figure.

The \textit{system plane} is responsible for keeping the overall system up to date to support reliable data queries and efficient model recommendations. 
Its functions are executed automatically whenever new labeled data arrives or a new trained model with a new data distribution is produced. 
Its key tasks are training the embedding model, training the clustering algorithm, updating the data store, verifying the uncertainty quantification of the clustering algorithm, and updating the index of training datasets associated with models in the model Zoo. 

\proj{} incorporates various built-in Autoencoder~\cite{bengio2014deep,wang2015unsupervised}, contrastive learning~\cite{chen2020simple}, and BYOL~\cite{grill2020bootstrap} embedding methods.
The user can select one of these algorithms, based on their application requirements and data specifications, or alternatively incorporate their own embedding model or algorithm into \dms{} by extending the embedding interface module. 
The \dms{} \textit{Training Embedding} module also supports tuning of hyper-parameters such as batch size and learning rate associated with an embedding module. 
Once an appropriate embedding algorithm and hyper-parameters are chosen, the training embedding module takes all the historical data from the data store as an input and trains the model for a user-defined number of epochs. 
Having trained the embedding model, the resulting embedding are used to train the clustering model, and the clustering module is then applied to assign each historical data item to the appropriate cluster. 

The embeddings and clusterings assigned to historical data can be updated in the data store periodically, triggered for example by spare resource availability~\cite{liu2021bftrainer} or by  uncertainty thresholds associated with the embedding or clustering models on new data. 
\proj{} monitors the uncertainty of the cluster assignment for each input dataset while servicing each user request.
If the input data are assigned to their respective clusters with high certainty, \dms{} simply performs the lookup operation from the data store and extracts the labeled data. 
In the case of high uncertainty, \proj{} triggers the training encoder and clustering function with the updated data from the data store. 
The intuition behind this step is that high uncertainty is due to staleness of the encoder and clustering model. 
We demonstrate the effectiveness of the uncertainty based trigger method for updating the embedding and clustering models in \figautoref{sec:uq-dms}. 

The platform backend is also responsible for managing the model Zoo, a task that tracking all historical models along with the distribution of their respective training data, as discussed in \secautoref{model_repo}. 

%% file: results.tex
\section{Experimental studies}\label{results}

We use two widely used light source data analysis applications to evaluate the performance of our \proj{} approach and implementation, applying each application to three different datasets. 
These applications are innately distinct from traditional industrial applications with respect to the type and volume of data and the resulting model complexity. 

\subsection{Benchmark Applications}
We consider two DNN models of different architecture and size:

\textbf{CookieNetAE}: The CookieBox detector~\cite{Audrey2019} is an angular array of 16 electron time-of-flight spectrometers. 
The X-ray shot photo-ionizes gas molecules in the interaction point, ejecting electrons. 
These electrons drift through a series of electrostatic potential plates in the 16 channels and are then detected by microchannel plates. This problem becomes difficult when we consider using a circularly polarized optical laser field in the interaction region and when the number of detected electrons is low. 
CookieNetAE is a deep neural network designed to estimate the energy-angle dependent probability density function of electron energies for all 16 channels. 
The network takes as input a 128$\times$128 image in which each of 128 rows corresponds to an empirical energy histogram, with 128 bins of 1~ev width, for a given CookieBox channel built after the time-energy mapping; it produces as output an image containing the probability density of electron energies in each channel~\cite{liu2021bridge}. 

\textbf{BraggNN}: X-ray characterization methods such as HEDM are used for designing and studying the properties of new materials. 
A single HEDM scan contains 1400--3600 frames, generated in 6--15 minutes today and an expected 50--100 seconds with APS-U~\cite{apsu}. 
Each scan contains multiple diffraction peaks, at different positions; analysis is then needed to determine the center of mass for each diffraction peak, with sub-pixel accuracy. 
BraggNN \cite{BraggNN-IUCrJ} is a deep learning model that has been shown to predict center of mass locations in a fast and robust manner compared to conventional pseudo-Voigt peak fitting. 
Recent studies have demonstrated that BraggNN can localize the center of mass 200$\times$ faster than conventional methods~\cite{BraggNN-IUCrJ}. 

\subsection{Benchmark Datasets}
We use three datasets of different sizes to evaluate how well \dms{} can store and extract data during different phases of the ML training cycle.

The \textbf{BraggPeaks} data are from 27 distinct X-ray diffraction microscopy experiments at APS. 
We pre-processed the data to split each 1440$\times$1440 pixel frame into a total of 1,868,228 distinct 15$\times$15 pixel patches, each containing a single Bragg peak.

The \textbf{CookieBox} data are from a computational simulation of the CookieBox detector. This simulation generates an image in which each of 128 rows correspond to an empirical energy histogram, with 128 bins of 1~ev width, of a given CookieBox channel built after the time-energy mapping, for a total of 
128$\times$128 8-bit integer pixels. 

\textbf{Tomography}: Synchrotron-based X-ray tomography is a noninvasive imaging technique that allows for reconstructing the internal structure of materials at spatial resolutions ranging from tens of micrometers to a few nanometers. 
Although CT experiments at synchrotrons can collect data at high spatial and temporal resolution, in-situ or dose-sensitive experiments require shorter exposure times to capture relevant dynamic phenomena with high temporal resolution or to avoid sample damage.
In such settings, a suitable DNN (e.g., TomoGAN~\cite{liu2020tomogan}) can be used to denoise low-dose images or to remove reconstruction artifacts from sparse views~\cite{liu2019deep,abeykoon2019scientific}.
In this dataset, each sample comprises 2048$\times$2048 16-bit integers. 

\subsection{Experimental Setup}
Our end-to-end workflow uses the Globus Flows service \cite{chard2019globus} to orchestrate funcX \cite{chard2020funcx} and Globus transfer \cite{foster1997globus} tasks, much as proposed by Liu et al.~\cite{liu2021bridge}. 
We use Globus Flows to define the end-to-end system flow, funcX as the serverless API for executing the user and system plane functions for optimal resource allocation, and Globus transfer to copy experimental data and model between experimental facility and compute cluster~\cite{vescovi2022linking}. 

\subsection{Impact of Storage System}
We first examine the impact of storage system on training speed (i.e., training time per epoch) and I/O time per iteration. 
We conducted experiments using Network File System (NFS) connected to the compute node with a 100 Gigabit Ethernet card. 
MongoDB also connected to the compute node through a 100 Gigabit Ethernet card. 
We evaluate MongoDB performance by using two data libraries, Blosc \cite{alted2017blosc} and Pickle \cite{Pickle}, to serialize the data to MongoDB format. 
\proj{} incorporates the PyTorch dataloader to speed up I/O reads for all experimental settings. 
The builtin PyTorch dataloader offers three abstractions: Dataset, Sampler, and DataLoader. 
Dataset returns a data item corresponding to a given index. 
Sampler creates random permutations of indices in the range of dataset length. 
Dataloader fetches one mini-batch worth of indices from the sampler, and adds these to the index queue. 
The worker processes of DataLoader consume these indices, and fetch data items from Dataset.
We extended the Dataloader functionality to read data efficiently from MongoDB with high parallelism (i.e., fetch using multiple clients) to hide the overhead of each fetch.


\begin{figure}
\centering    
\subfloat[Impact of Batch size]{\label{subfig:tomo_pickle_h5_batchsize}\includegraphics[width=0.5\columnwidth]{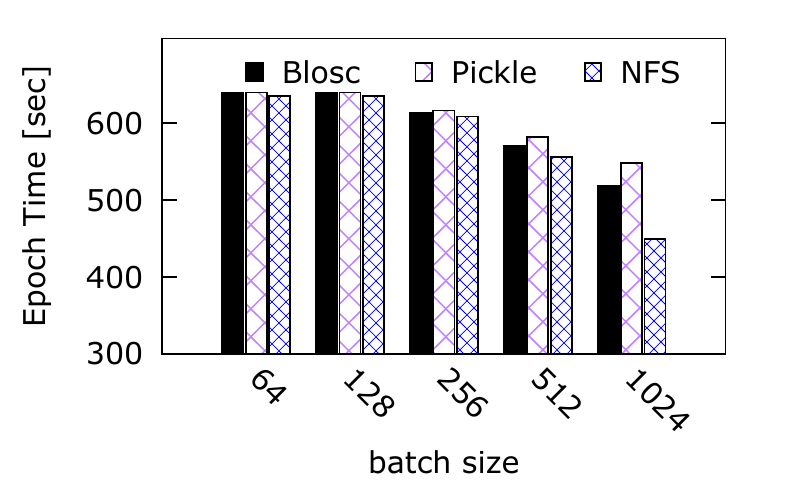}}
\subfloat[Impact of \# Workers ]{\label{subfig:tomo_pickle_h5_workers}\includegraphics[width=0.5\columnwidth]{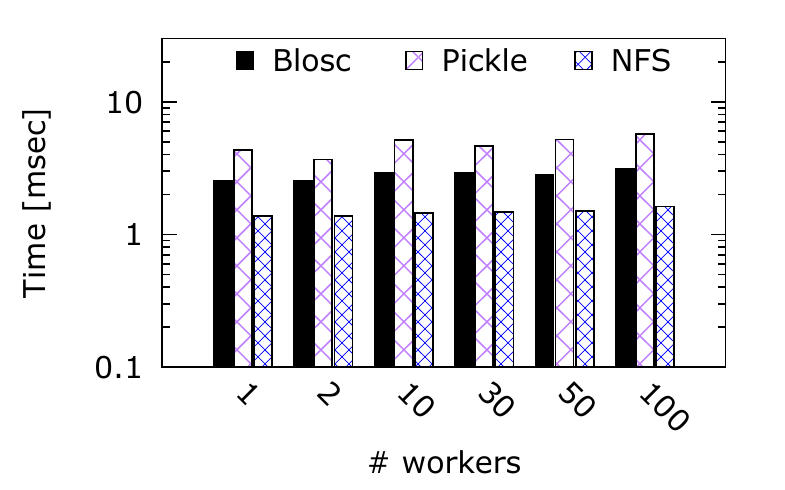}}
\caption{Reading Tomography data from MongoDB hosted remotely using Pickle data encoding and H5, for varying batch size and number of workers.}
\label{fig:tomo_pickle_h5_remote}
\end{figure}

\begin{figure}
    \centering    
         \subfloat[Impact of Batch size]{\label{subfig:cookiebox_pickle_h5_batchsize}\includegraphics[width=0.5\columnwidth]{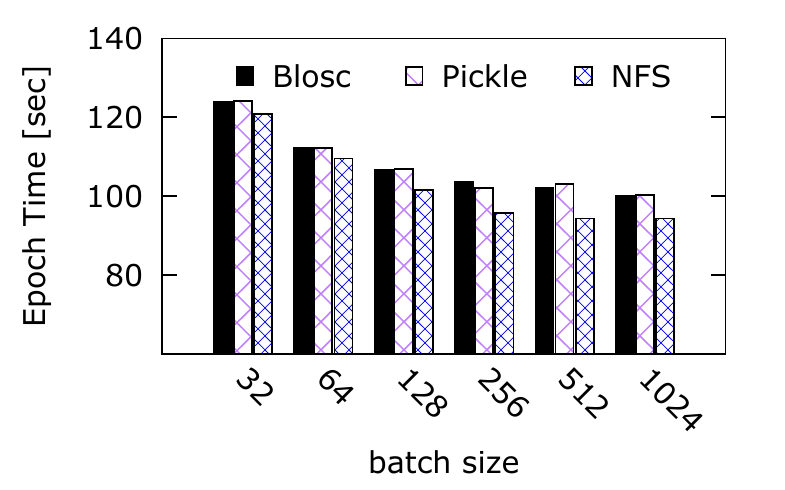}}
         \subfloat[Impact of \# Workers ]{\label{subfig:cookiebox_pickle_h5_workers}\includegraphics[width=0.5\columnwidth]{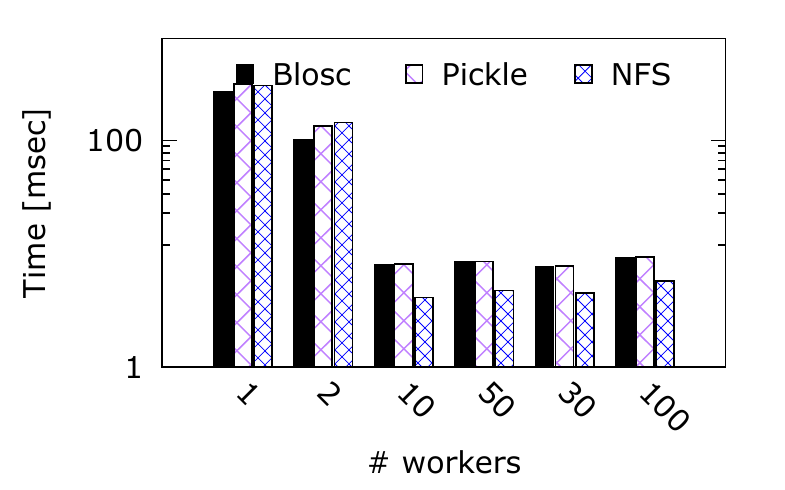}}
    \caption{Reading CookieBox data from MongoDB hosted remotely using Pickle data encoding and H5, for varying batch size and number of workers.}
    \label{fig:cookiebox_pickle_h5_remote}
\end{figure}

\begin{figure}
    \centering    
         \subfloat[Impact of Batch size]{\label{subfig:bragg_pickle_h5_batchsize}\includegraphics[width=0.5\columnwidth]{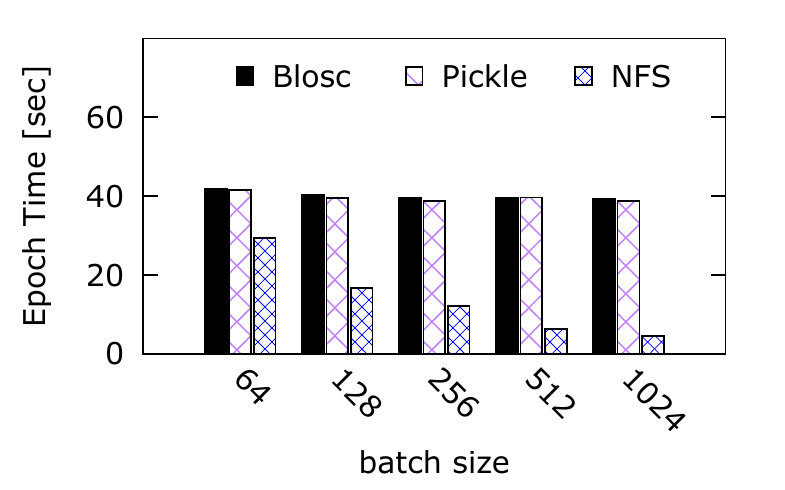}}
         \subfloat[Impact of \# Workers ]{\label{subfig:bragg_pickle_h5_workers}\includegraphics[width=0.5\columnwidth]{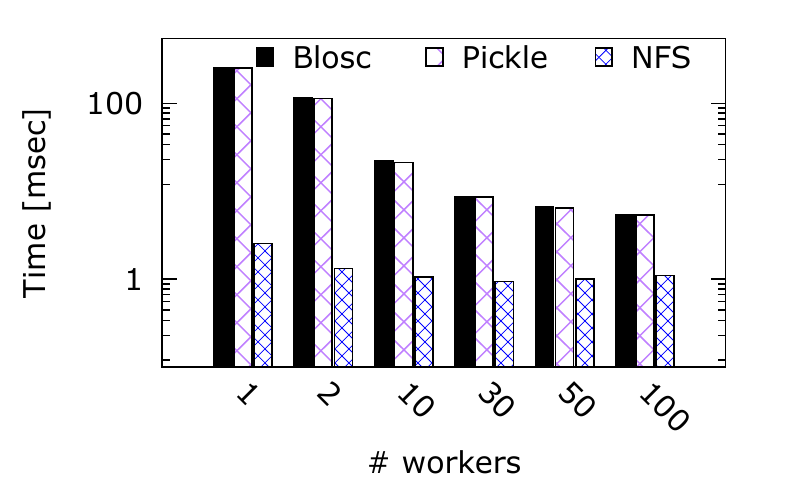}}
    \caption{Reading Bragg data from MongoDB hosted remotely using Pickle data encoding and H5, for varying batch size and number of workers.}
    \label{fig:bragg_pick_h5_remote}
\end{figure}

\figautoref{fig:tomo_pickle_h5_remote}--\ref{fig:bragg_pick_h5_remote} show the per-epoch training times and per-iteration I/O times for the Tomography, CookieBox, and BraggPeaks datasets, respectively. 
Each left-hand subfigure shows the training time per epoch as a function of batch size with a fixed number of 50 I/O threads; 
In the case of the Tomography and CookieBox datasets (\figautoref{subfig:tomo_pickle_h5_batchsize} and \figautoref{subfig:cookiebox_pickle_h5_batchsize}, respectively) we can a trend similar to that seen in any conventional ML workload, with the epoch time inversely proportional to batch size. 
We note that different storage configurations all achieve similar performance, except for the case of batch size 1024 in \figautoref{subfig:tomo_pickle_h5_batchsize}, where NFS outperforms both Blosc and Pickle. 
This is due to the overhead incurred by deserialization of the training data from Blosc and Pickle formats.

Each right-hand subfigure shows I/O time per iteration for varying numbers of concurrent threads reading data from storage
(shown on the x-axis as \# workers) a fixed batch size of 512. 
We can observe the impact of deserialization in I/O time as a function of workers concurrently reading training data per iteration as well, which is demonstrated in \figautoref{subfig:tomo_pickle_h5_workers} and \figautoref{subfig:cookiebox_pickle_h5_workers}, where NFS also outperforms both Blosc and Pickle. 
We attribute the lack of improvement in the I/O time of NFS impacting the epoch time in smaller batch sizes (i.e., less than 1024) to two reasons.
First, the improvement is only a few hundred milliseconds compared to the computation time per iteration, which is in the order of tens of seconds. 
Second, the PyTorch Dataloader prefetching allows the data to be read into RAM in parallel with computation. 
We conclude that for the Tomography and CookieBox datasets, the choice of storage system does not significantly impact training performance.

We see similar trends for the Bragg dataset in \figautoref{subfig:bragg_pickle_h5_batchsize}.
However, in this case direct reading from NFS improves the training time per epoch significantly when compared to Blosc and Pickle. 
This behaviour variation is due to the large model and dataset sizes of the Bragg dataset, which is latency-bound due to its size. 
Thus, there is no significant variation in NFS I/O time for different numbers of worker threads reading the data. 
On the other hand, Blosc and Pickle demonstrate shorter I/O times when more workers are used for prefetching, denoting that the cost of fetching data from MongoDB is higher than that of fetching directly from files, but that this cost can be mitigated byusing higher parallelism. 

We conclude from these experiments that while the use of MongoDB-based storage has no significant impact on training time \reviews{for the Tomography and CookieBox datasets} compared with reading directly from NFS, the rich features provided by MongoDB (e.g., inserting, updating, deleting, querying) can significantly reduce the service and data updating overheads (compared to operating and managing files directly) when updating the \dms{} backend. 
\reviews{In the case of Bragg data we propose to store the historical data in the MongoDB for the ease of data management. However, prefetching the training data from MongoDB to NFS (or even a local SSD) in the beginning would help improve the overall training time.}
  


\subsection{Data Service Validation}

We use the BraggNN model and BraggPeaks data to evaluate the effectiveness of the \dms{} in identifying similar data for model training.
Specifically, we select an HEDM experimental dataset not in BraggPeaks, which we denote here as $B_R$, identify a holdout set $B_H\in B_R$, and define a threshold $T$. Then we construct $B_{O}$ as follows: For each element $b\in B_R\setminus B_H$, we identify the image $p$ in BraggPeaks that is closest to $b$ in embedding space, and if $|b - p| < T$, retrieve the label $l(p)$ and add $\{p, l(p)\}$ to $B_{O}$; if $|b - p|\ge T$, we use the pseudo-Voigt code to compute a label $v(b)$ for $b$, and add $\{b, v(b)\}$ to $B_{O}$.

\begin{figure}[htb]
\centering    
\subfloat[Conventional method.]{\label{subfig:bragg_labeled_data}\includegraphics[width=0.95\columnwidth]{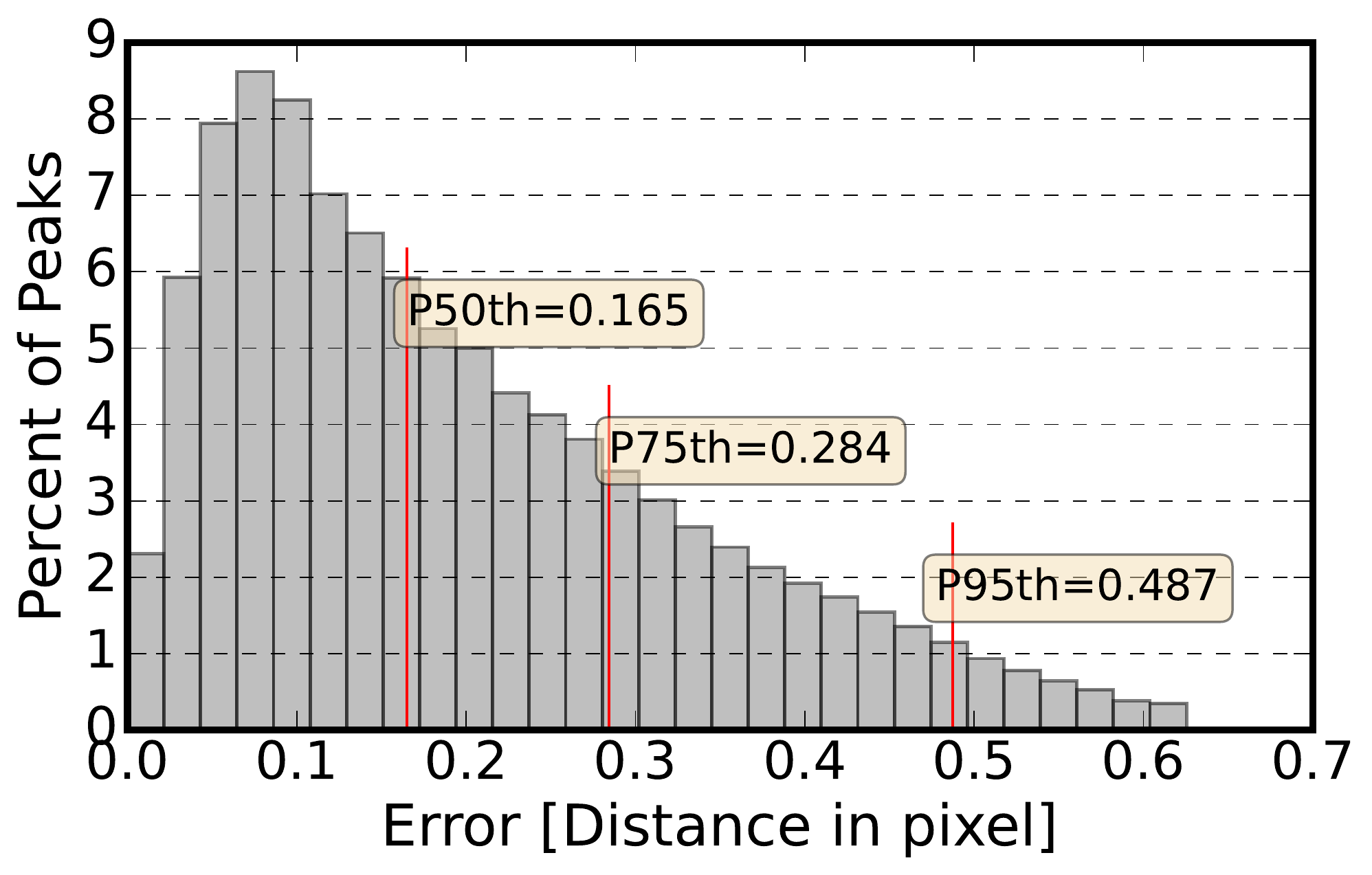}}\hfill
\subfloat[Proposed \dms{}.]{\label{subfig:bragg_pseudo_labeled_data}\includegraphics[width=0.95\columnwidth]{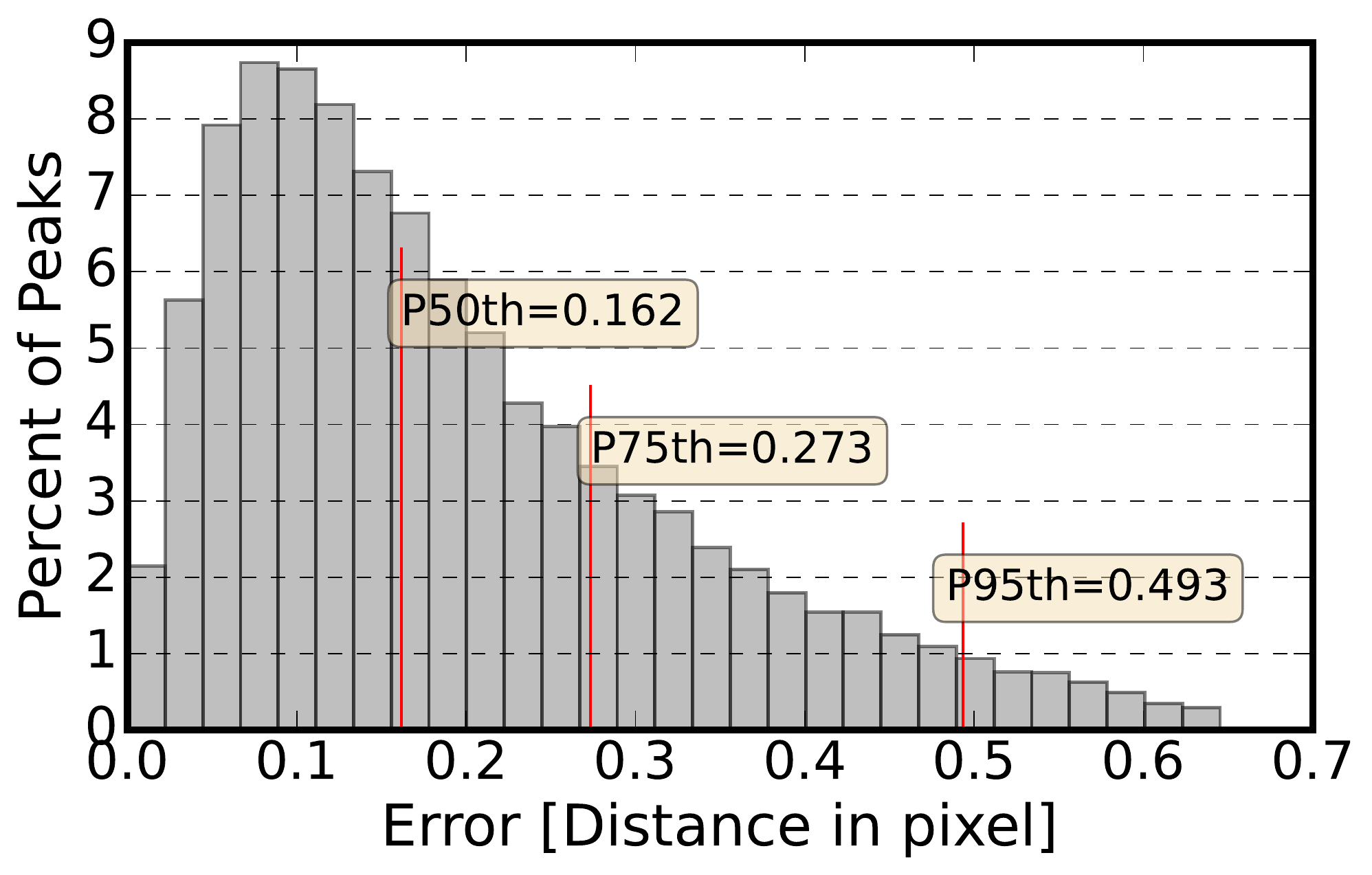}}
\caption{Prediction error distribution of BraggNN trained with a conventionally labeled dataset vs.\ a historical dataset retrieved by using \dms{}.}
\label{fig:pseudo-labeling-braggnn}
\end{figure}

We then train the BraggNN model on $B_R\setminus B_H$ and $B_O$ to obtain two trained models, and compare the prediction errors of the two resulting models on $B_H$. 
We see in \figautoref{fig:pseudo-labeling-braggnn} that the two models performed equally well.
However, the times spent on labeling were significantly different; the conventional method needs nearly an hour while \dms{} took less than a minute. 
We examine end-to-end performance in terms of labeling time and training time in more detail in \secautoref{case_study}. 


\subsection{Model Service Validation}\label{sec:MSval}
We evaluate \mms{} by using a test dataset that is new to \proj{} and labelled it by using conventional methods.
For each model in the \mms{} Zoo, we:
(a) applied the model to each element in the test dataset, and computed the mean prediction error across all elements; and
(b) computed, as a measure of dataset similarity, the distance between the data distributions for the test dataset and the dataset used to train the model.
We then use the results of these computations to  
study the relationship between prediction error and dataset similarity.

\begin{figure}[htb]
\centering    
\subfloat{\label{subfig:acuracy_dist_1}\includegraphics[width=0.5\columnwidth]{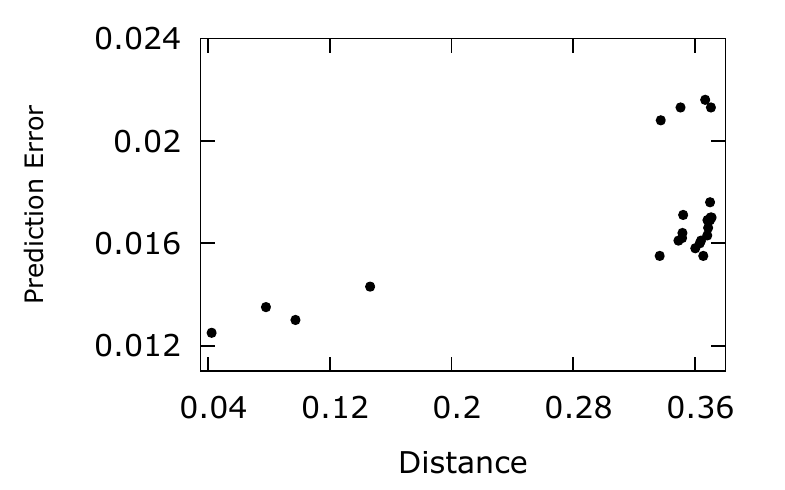}}
\subfloat{\label{subfig:acuracy_dist_2}\includegraphics[width=0.5\columnwidth]{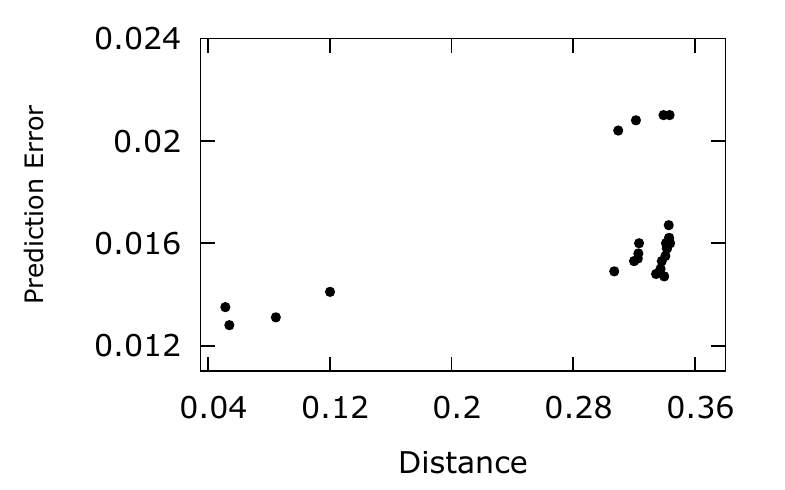}} \\ 
\subfloat{\label{subfig:acuracy_dist_3}\includegraphics[width=0.5\columnwidth]{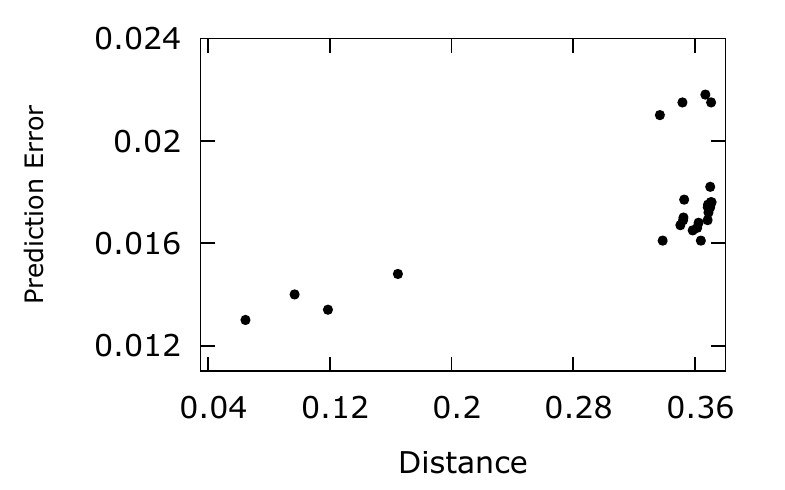}}
\subfloat{\label{subfig:acuracy_dist_4}\includegraphics[width=0.5\columnwidth]{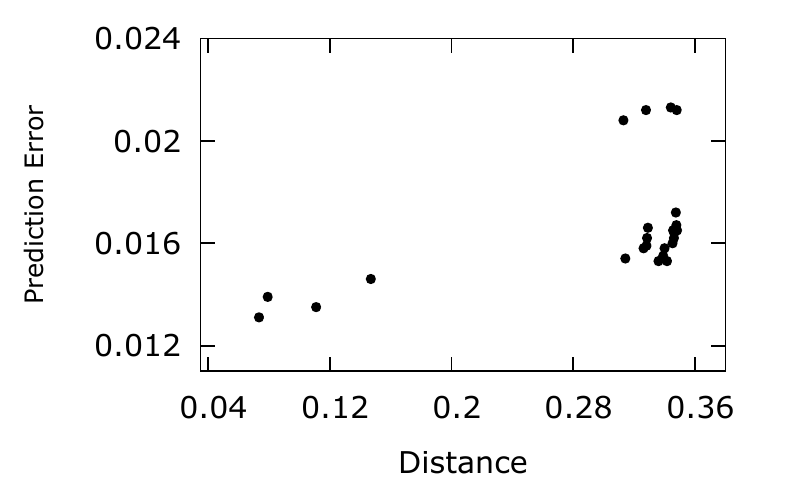}}
\caption{Prediction error vs.\ dataset distance for BraggNN on four different datasets.}
\label{fig:braggacuracy_dist}
\end{figure}

\begin{figure}[htb]
\centering    
\subfloat{\includegraphics[width=0.5\columnwidth]{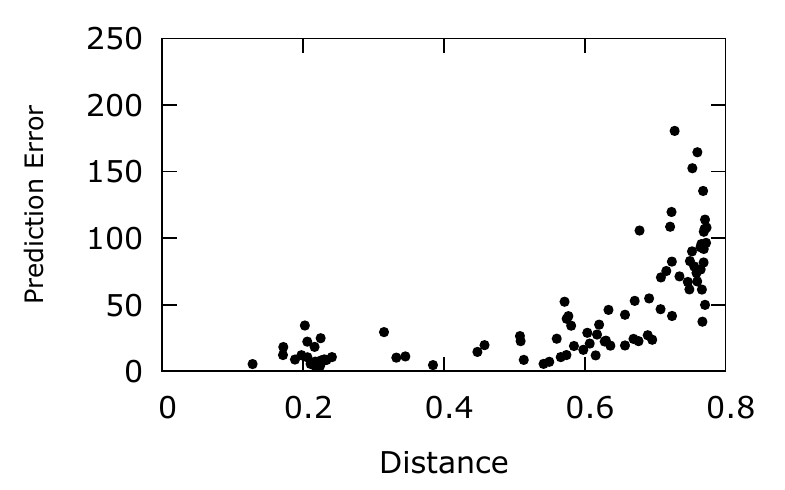}}
\subfloat{\includegraphics[width=0.5\columnwidth]{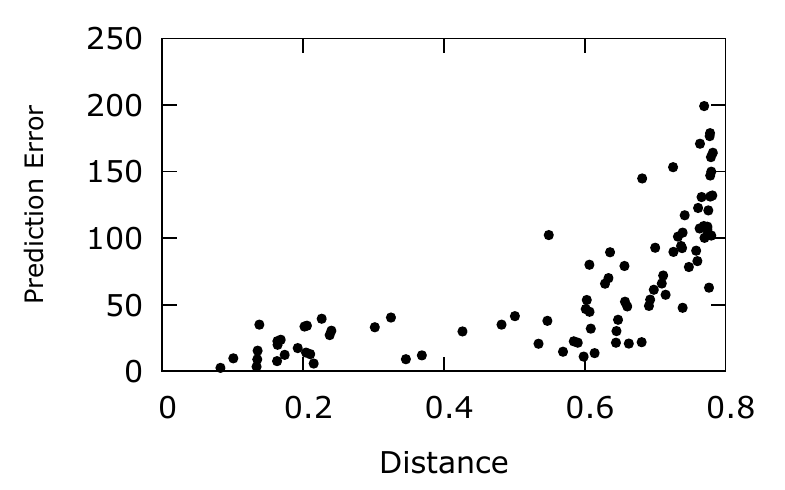}}\\
\subfloat{\includegraphics[width=0.5\columnwidth]{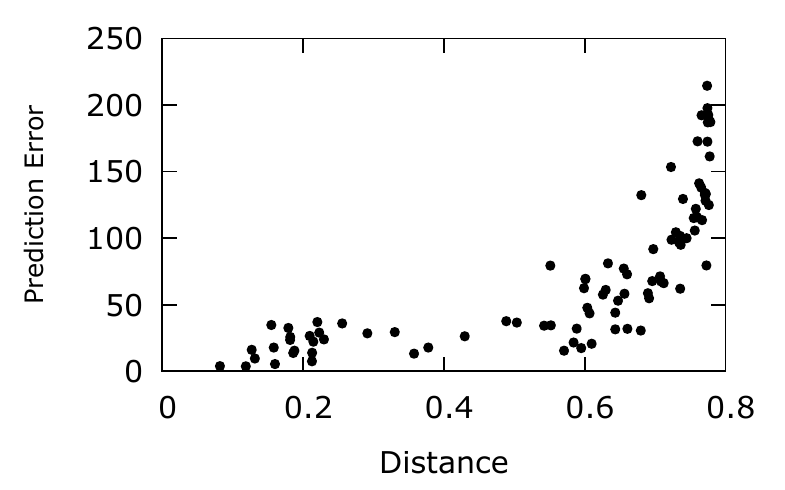}}
\subfloat{\includegraphics[width=0.5\columnwidth]{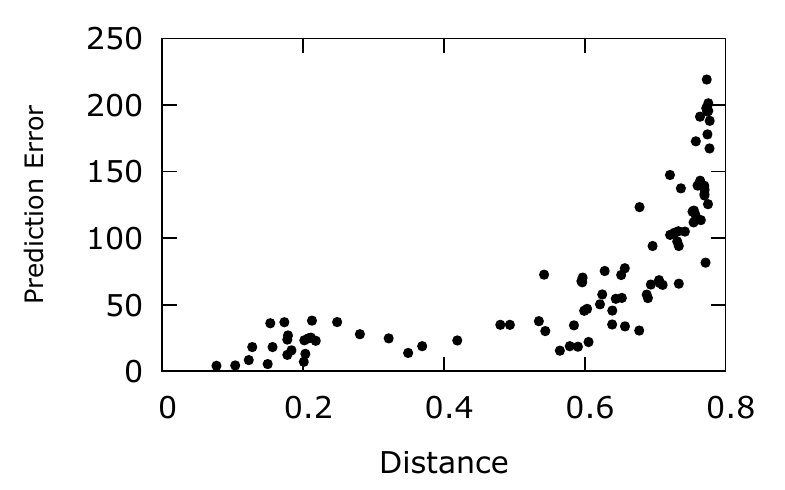}}
\caption{Prediction error vs.\ dataset distance for CookieNetAE on four different datasets.}
\label{fig:Cookieboxacuracy_dist}
\end{figure}

\figautoref{fig:braggacuracy_dist} and \figautoref{fig:Cookieboxacuracy_dist} show scatter plots of model prediction error vs.\ JSD distance between the model's training dataset and the test dataset,
for the BraggNN and CookieNetAE datasets, respectively.
In each figure, the y-axis is the prediction error of the model, and the x-axis is the JSD distance between the dataset used for training the model and the input test dataset. 
Lower values on the y-axis reflect higher model accuracy, while points further from the x-axis indicate more different datasets. 
Therefore, points closer to the x-axis have lower prediction error (i.e., models with high accuracy), while points closer to the y-axis have smaller JSD (i.e., the dataset has high similarity). 
An ideal model for fine-tuning would have both low JSD distance and low prediction error. 
For \mms{} to be able to produce an ideal model, the relationship between model performance and dataset distance need to be monotonic (possibly with some uncertainty). \reviews{This behavior is dependent on the distribution of data used for training the model. For example, in the case of \figautoref{fig:Cookieboxacuracy_dist}, we observe generally monotone behavior because the data used for training the models changes slightly over time, resulting in a slight degradation of models over time. On the other hand, in the case of \figautoref{fig:braggacuracy_dist} the variation in the data is bimodal (i.e, in the early phase of the experiment the configurations are similar, and hence have similar data distributions, while in the second phase, we see a different data distribution based on a new configuration).}

\reviews{Even without the perfect monotonic behaviour in \figautoref{fig:braggacuracy_dist} and \figautoref{fig:Cookieboxacuracy_dist}} our results demonstrates that \mms{} can efficiently meet these requirements and find the best model for fine tuning.  
We see in both applications that prediction error and distance are positively correlated: the model with the smaller distance generally has the lower prediction error. 
Therefore, we conclude that \mms{}'s dataset JSD-based model ranking is effective for fine-tuning models for robust model training.

\begin{figure}[htb]
\centering
\includegraphics[width=\columnwidth]{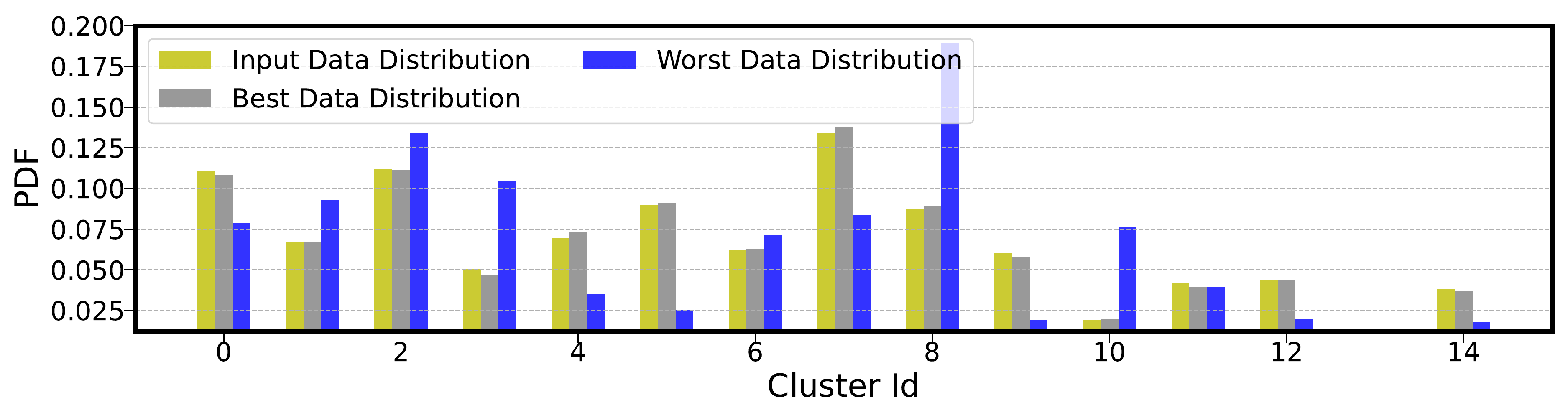}
\caption{Data distribution comparison of input dataset vs.\ training dataset distribution of best ranked model and worst ranked model.}
\label{fig:Data-distribution-compare-BraggNN}
\end{figure}

To further verify and interpret the ranking of models based on the dataset distribution distance of the input data with the training dataset, we demonstrate the comparison of best ranked and worst ranked model data distribution with the input data distribution. 
\figautoref{fig:Data-distribution-compare-BraggNN} illustrates the data distribution comparison where the x-axis shows the cluster ID of 15 clusters of the Bragg dataset while y-axis shows the distribution of each cluster in the input dataset, training data of best ranked model and training data of worst ranked model represented by green, gray, and blue bars respectively. 
We see that both the input data and the best model follow the same distribution, while the worst ranked model follows quite a different trend. 
We note that there is no data points assigned to cluster 13 for all three model data distributions.

These results complement those discussed in the previous section, where we observed that the best ranked model has the smallest JSD value. 
Next, we evaluate the enhancements in training speed that result from the use of the fine tuning approach with the best, median, and worst recommended model vs.\ a model trained from randomly initialized model parameters (i.e., re-trained from scratch).

\subsection{Rapid DNN Training with \proj{}}
We  evaluate the effectiveness of \proj{} by comparing the end-to-end time required to update a ML model with a given (unlabeled) dataset. 
In these experiments we demonstrate how \mms{} can recommend the best ML model from the Zoo for fine-tuning that leads to significant improvements in training time. 
We compare the time required to train the recommended model to convergence with that required for the median ranked model, the worst ranked model, and a model with randomly initialized parameters.

\begin{figure}[htb]
\centering    
\subfloat{\label{subfig:subfig:cb_e2e_95}\includegraphics[width=\columnwidth]{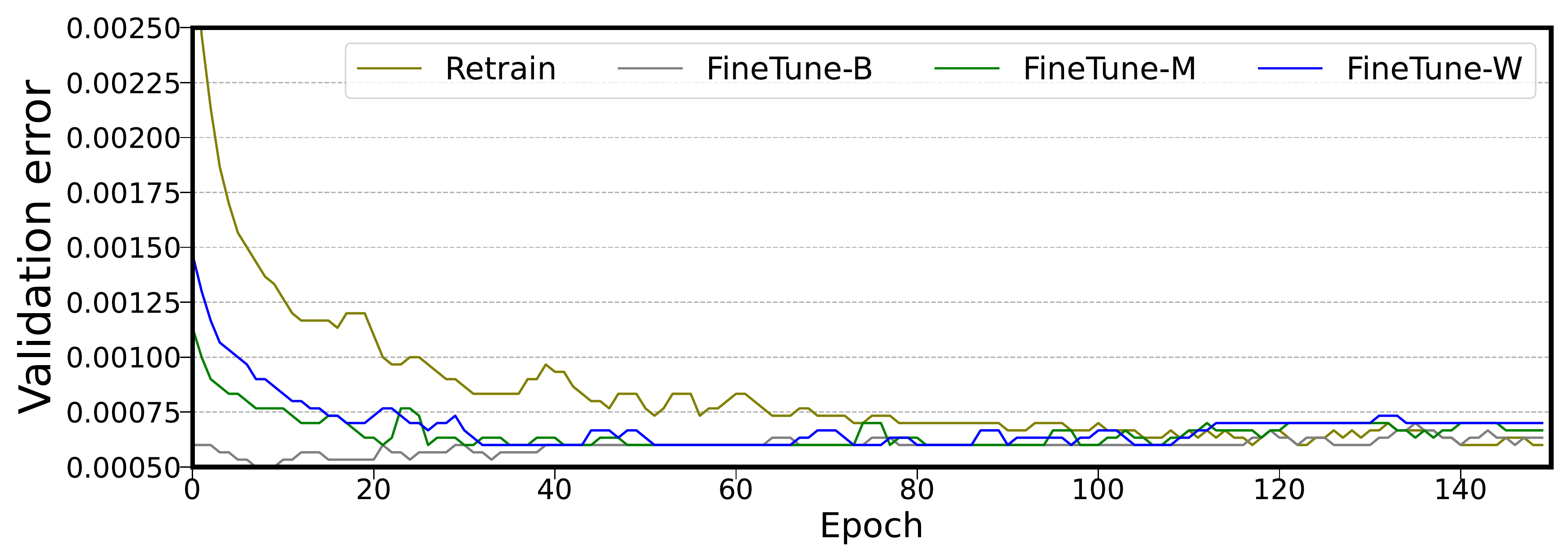}}\hfill
\subfloat{\label{subfig:subfig:cb_e2e_97}\includegraphics[width=\columnwidth]{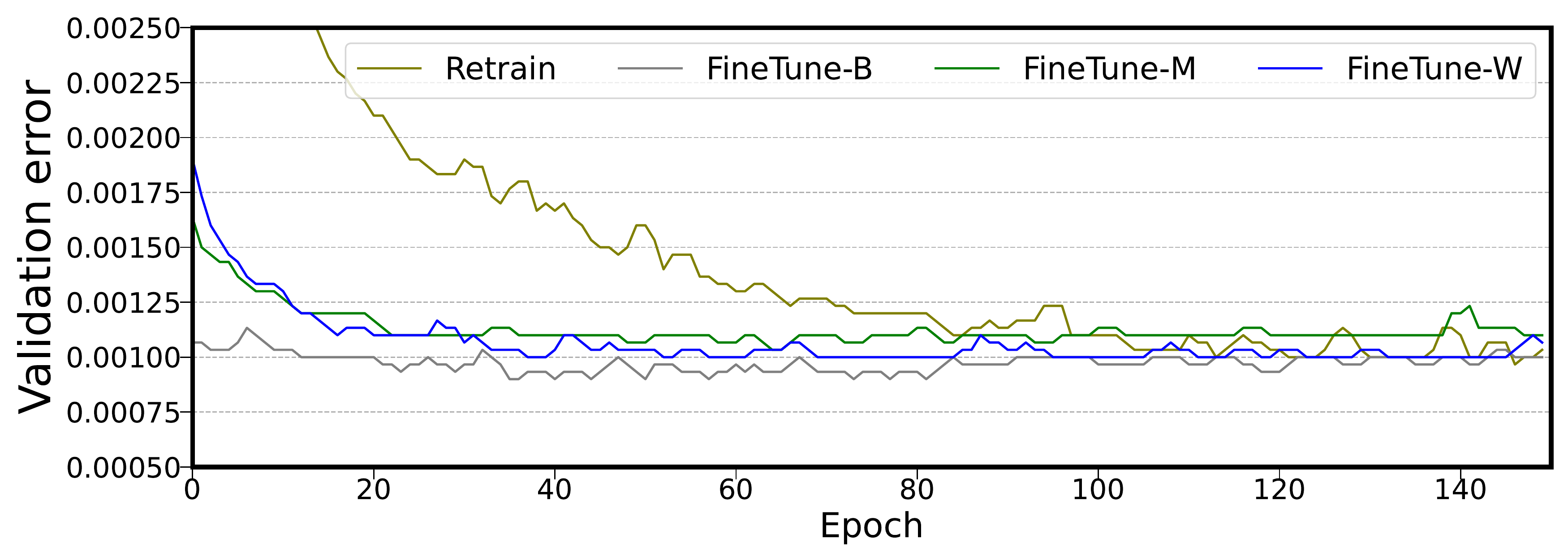}}\hfill
\subfloat{\label{subfig:subfig:cb_e2e_98}\includegraphics[width=\columnwidth]{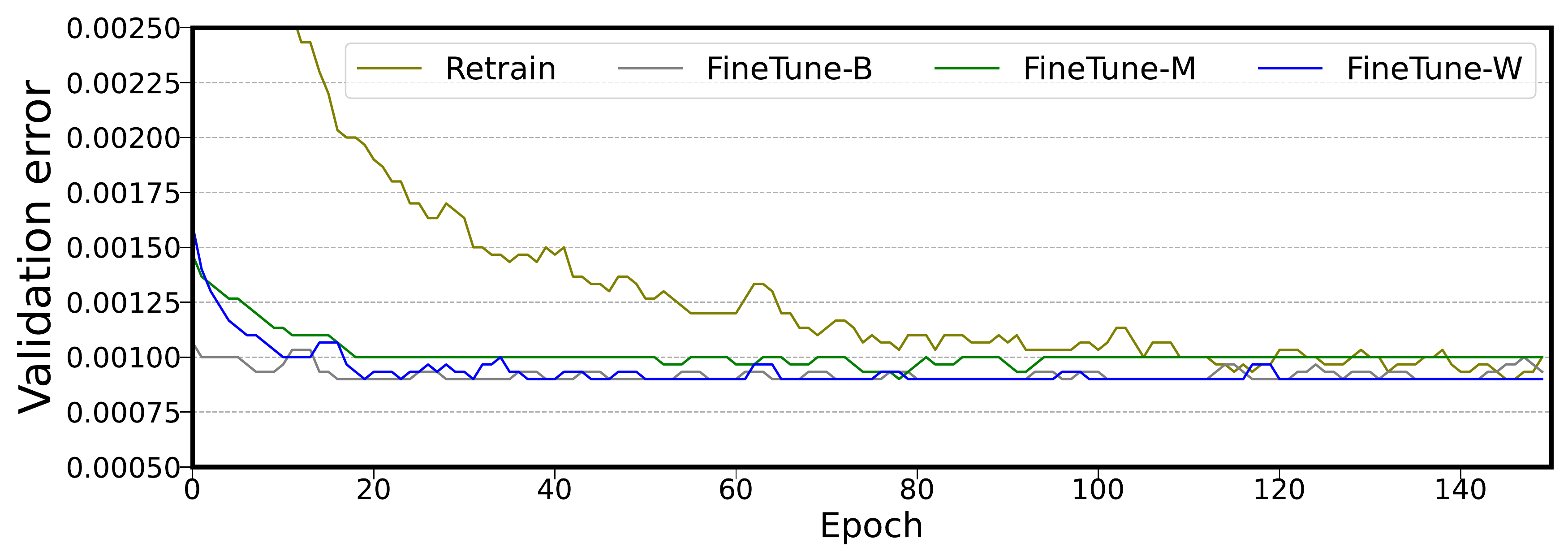}}\hfill
\subfloat{\label{subfig:subfig:cb_e2e_99}\includegraphics[width=\columnwidth]{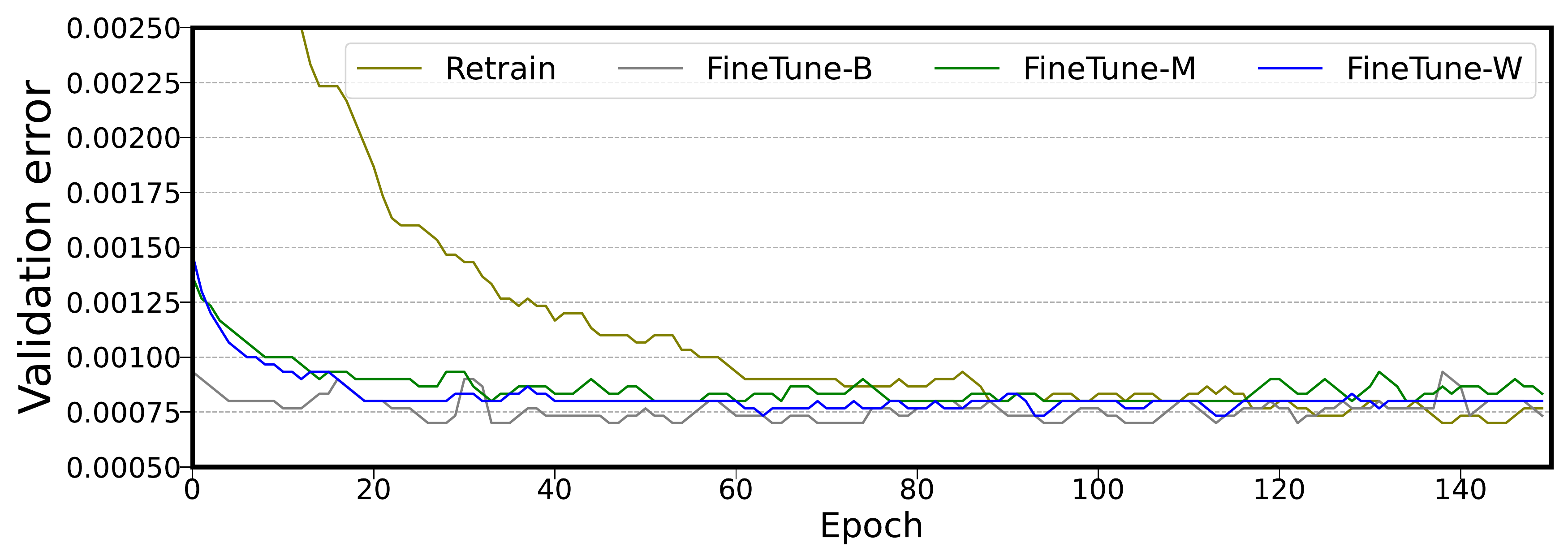}}\hfill
\caption{Learning curve of different training strategies for CookieNetAE on four different datasets.}
\label{fig:e2e_CookieNetAE}
\end{figure}

\begin{figure}[t]
\centering    
\subfloat{\label{subfig:subfig:bragg_e2e_661}\includegraphics[width=\columnwidth]{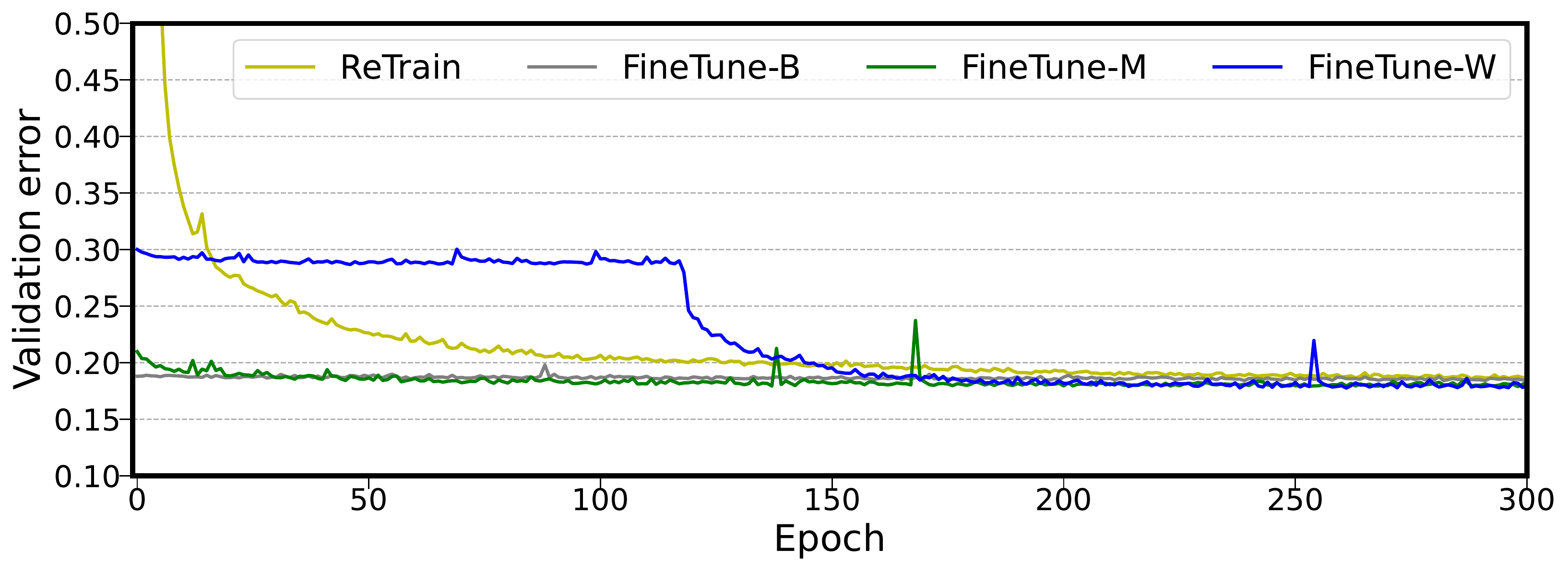}}\hfill
\subfloat{\label{subfig:subfig:bragg_e2e_465}\includegraphics[width=\columnwidth]{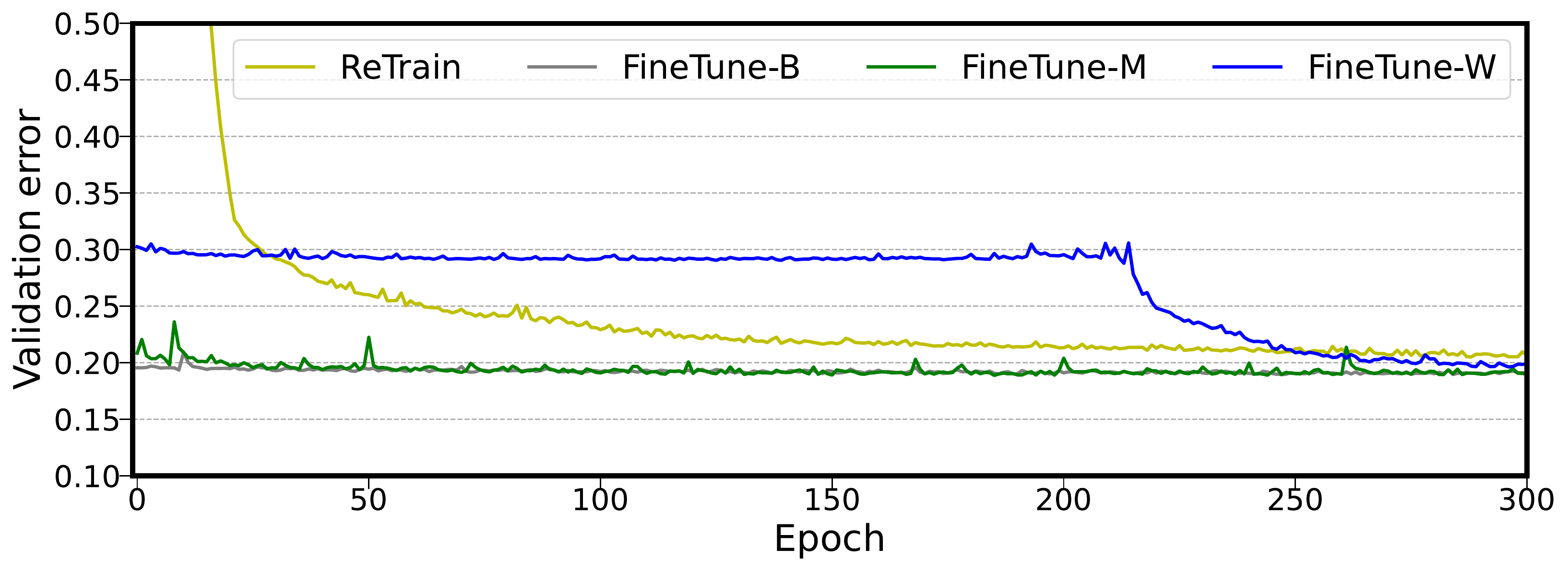}}\hfill
\subfloat{\label{subfig:subfig:bragg_e2e_468}\includegraphics[width=\columnwidth]{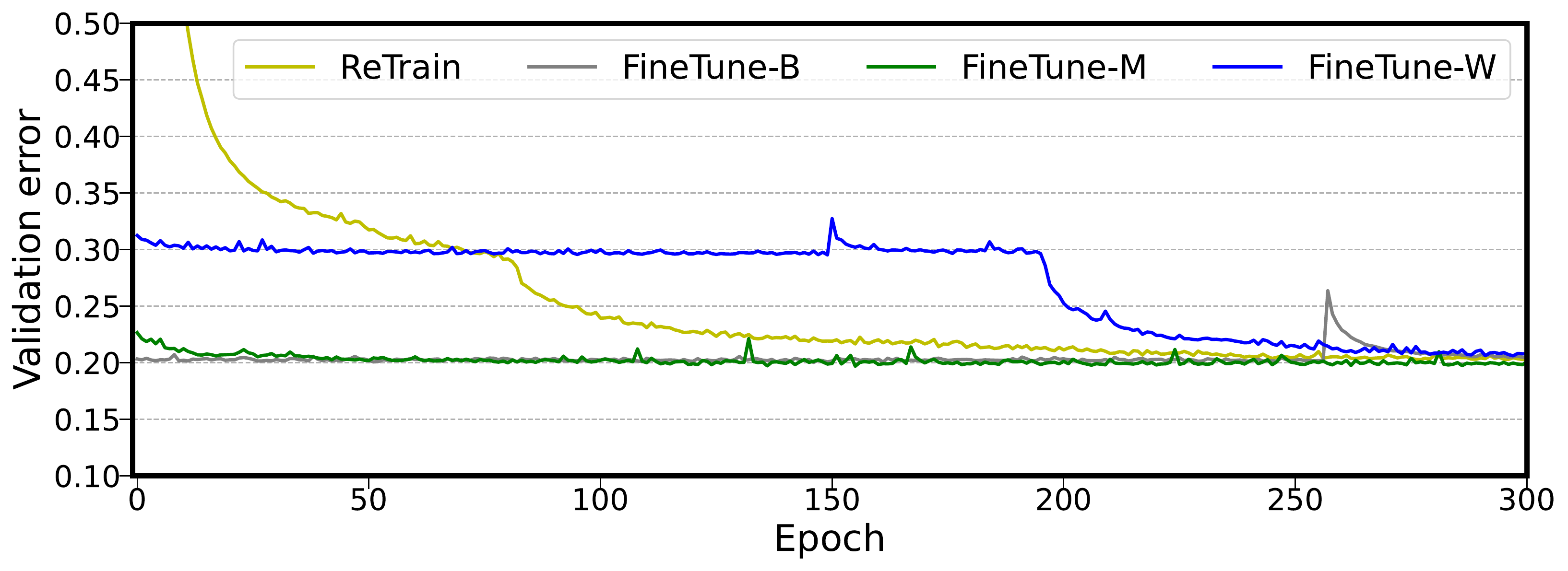}}\hfill
\subfloat{\label{subfig:subfig:bragg_e2e_471}\includegraphics[width=\columnwidth]{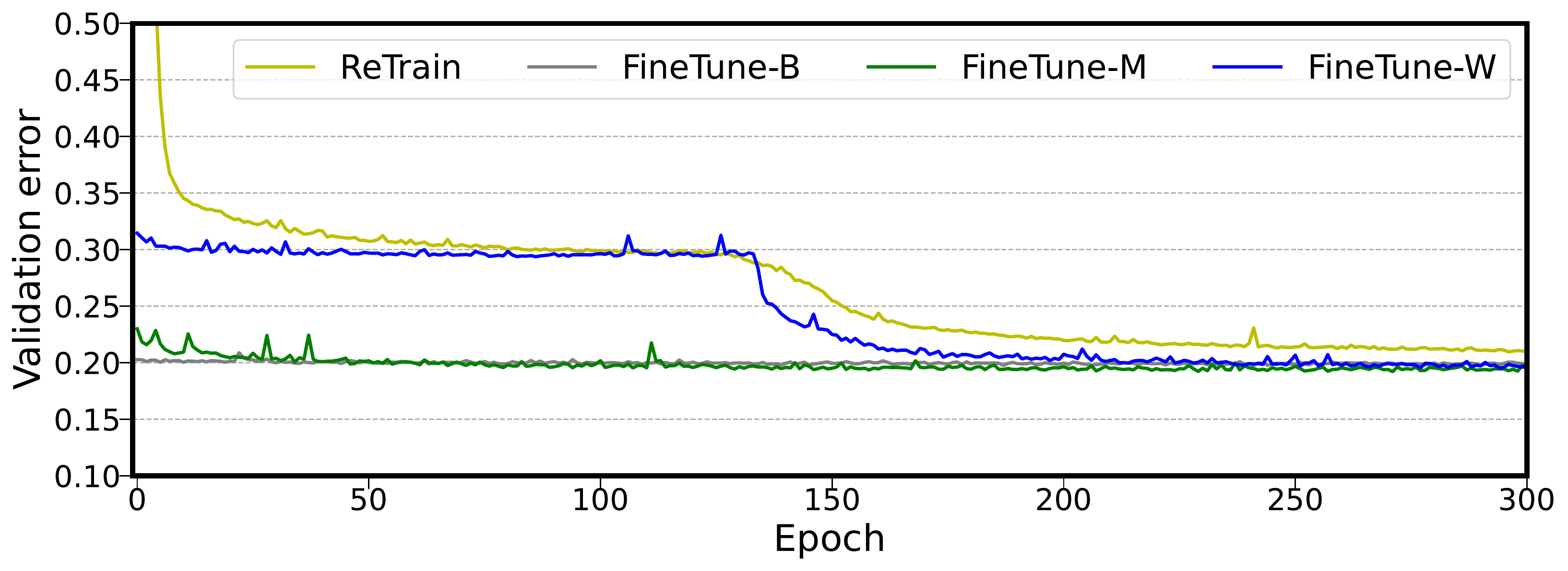}}\hfill
\caption{Learning curve of different training strategies for BraggNN on four different datasets.}
\label{fig:e2e_BraggNN}
\end{figure}

Figs.~\ref{fig:e2e_CookieNetAE} and \ref{fig:e2e_BraggNN} compare validation loss as a function of number of training epochs for a model trained from scratch (\textit{Retrain}) and for the Best, Median, and Worst model recommendations by \proj{} (denoted as \textit{FineTune-B}, \textit{FineTune-M}, \textit{FineTune-W}, respectively) for the BraggNN and CooikeNetAE datasets, respectively. 
We see that for both \textit{BraggNN} and \textit{CooikeNetAE} the optimal model recommendation by \proj{} for fine tuning (\textit{FineTune-B}) always reaches convergence within the first few epochs, performing consistently better than the median and worst model recommendations. 
Training from scratch (\textit{Retrain}) always converges slowly due to randomly initialized model parameters. \reviews{We note that in the case of \textit{CooikeNetAE} (in \figautoref{fig:e2e_CookieNetAE}), \textit{FineTune-M} and \textit{FineTune-W} take the same number of epochs to converge. This is due to the training data distribution similarity of both these models. This behavior is also observed in \textit{BraggNN} (\figautoref{fig:e2e_BraggNN}) where \textit{FineTune-B} and \textit{FineTune-M} demonstrate similar performance. We would expect to see more pronounced differences in convergence times in situations with greater diversity in training datasets, as might arise, for example, when using \dms{} at a beamline with a greater diversity of training configurations and samples. Nevertheless,}
these results demonstrate how \dms{} allows \proj{} to calculate effectively the data distribution of the input dataset, query labeled data to reduce the data labeling time, and finally find the best model for fine tuning using \mms{}, instead of tuning a model from scratch.


\subsection{Case Study: Retraining the BraggNN Model}\label{case_study}
Now that we have validated that \proj{} can accurately label data and recommend models, we want to examine whether and how its use can also improve training times when compared to a legacy approach in which data are labeled using conventional methods and the ML model is always re-trained from scratch. 
To this end, we conduct a case study involving retraining of the BraggNN model.

Specifically, we compare the training times observed when two different workflows (\proj{} and legacy) are applied in an HEDM experiment that has generated a series of datasets at successive times, indexed as 0,~1,~\ldots,~145. 
We assume that while processing dataset 21, we have determined that the ML model is not longer performing appropriately, and thus retraining is required before dataset 22. 
In the legacy workflow, all data to be used by experiment 22 must be labeled by using the compute-intensive pseudo-Voigt code, after which the BraggNN model must be trained from scatch.
In the \proj{} workflow, on the other hand, \proj{} can instead use the \dms{} to retrieve labels for data similar to the step 22 data, and then, furthermore, fine tune a historical model recommended by \mms{} instead of training BraggNN from scratch.

\begin{figure}[t]
\centering    
    \subfloat[Label vs.\  train]{\label{subfig:bragg_e2e_label_vs_train}\includegraphics[width=\columnwidth]{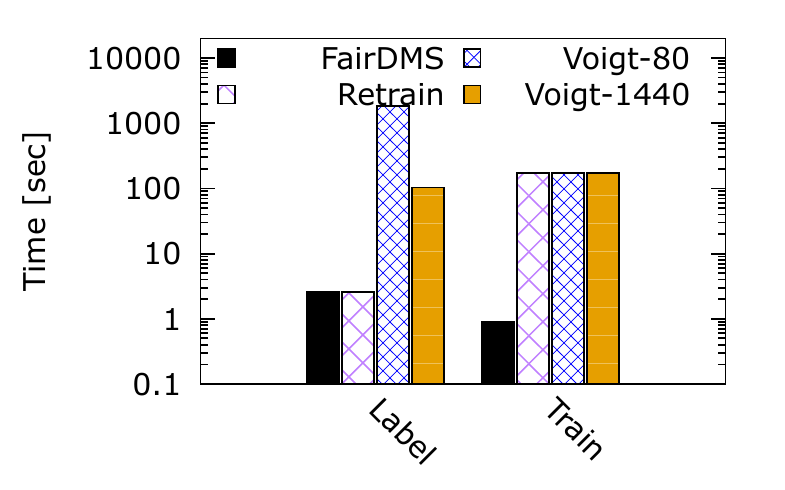}}
    
    \subfloat[End-to-end training time]{\label{subfig:bragg_e2e_time}\includegraphics[width=\columnwidth]{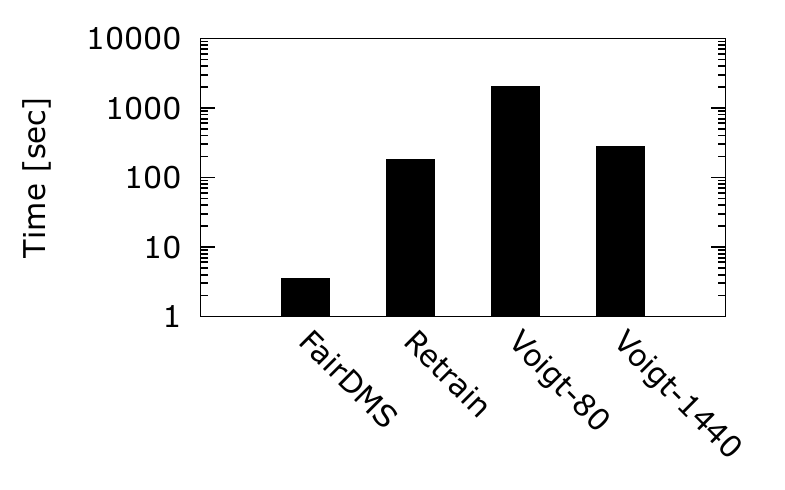}} 
\caption{Comparison of data labeling time, model training time, and end-to-end training time in the BraggNN case study.}
\label{fig:e2e_time_bragg}
\end{figure}

We compare the performance achieved by \proj{} with that of three alternative methods: 
1) Retrain: Here we use \dms{} for pseudo-labeling, but always train a model from scratch rather than using \mms{} to identify a previously trained model for fine tuning.
This experiment allows us to quantify the improvements achieved by \dms{} alone. 
2) Voigt-80: Here we employ the MIDAS~\cite{sharma2016midas} pseudo-Voigt code to label the experimental data, running on a single 80-core workstation (the most accessible solution) and training from scratch. 
This is the method that would conventionally be used in the absence of \proj{}; it serves as our baseline. 
3) Voigt-1440: Here, we employ the conventional method, but run the MIDAS code on an 18-node cluster, with a total of 1440 CPU cores---the highest possible parallelism supported by  MIDAS. This experiment provides a best case comparison for the conventional method.

\figautoref{subfig:bragg_e2e_label_vs_train} compares the  labeling times and training times for the four cases, with the y-axis giving time in seconds using a log scale. 
\proj{} and Retrain both take advantage of \dms{}, and hence show the lowest labeling times as they can reuse previously computed labels.
Voigt-80 takes the longest to finish labeling, even with 80 cores. 
Voigt-1440 labels faster than Voigt-80, thanks to its 18$\times$ greater computing power, but is still slower than the \dms{}-based cases. 
For training (performed in all cases one NVIDIA V100 GPU), 
\proj{}'s fine tuning approach allows it to converge 200$\times$ faster than the other three methods. 
This demonstrates that \mms{} can achieve the same model performance 200$\times$ faster compared to existing methods.

\figautoref{subfig:bragg_e2e_time} compares the total time taken by each method from when data arrives at the compute facility to when a trained model is returned to the user:
the sum of the data labeling and model training times. 
Overall, the Voigt-80 conventional baseline performs the worst of the four methods: almost 600$\times$ worse than \proj{}. 
This is due to both its slow labeling and the fact that it must train from scratch. 
\proj{} achieves 92$\times$ speedup even relative to Voigt-1440, which uses 1440 cores concurrently for labeling and the same amount of training resources. 
\proj{} is 58$\times$ faster than Retrain.

\subsection{Uncertainty Quantification of Learned Representations}\label{sec:uq-dms}

In this final study we demonstrate how \proj{} can use uncertainty quantification methods to determine when to retrain a model.
We use a sequence of datasets from HEDM experiments, numbered 0 to 35.
We first train our embedding and clustering models (with 15 clusters) by using the first five datasets.
Then, for successive datasets, we calculate the certainty of the clustering algorithm by using fuzzy k-means~\cite{fuzzy-k-means}, with certainty calculated in terms of the percentage of the dataset that are assigned to their respective cluster with at least 50\% confidence.

\begin{figure}
\centering    
\label{subfig:bragg_uq}\includegraphics[width=1.0\columnwidth,trim=3.5mm 4mm 3.5mm 4mm,clip]{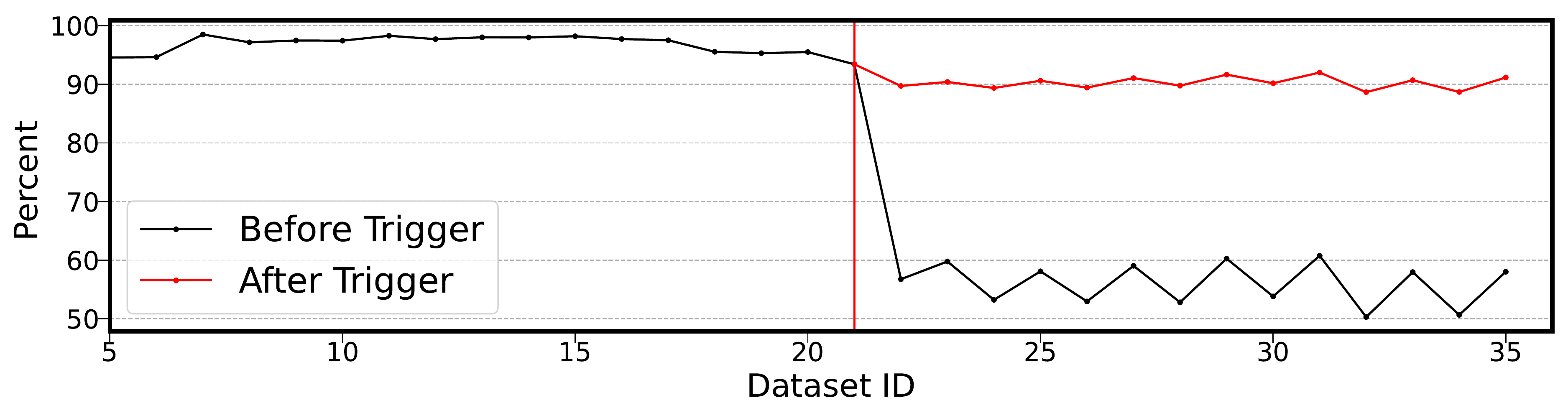} \\
\caption{Uncertainty quantification for the embedding model and clustering model in \dms{}, when run without and with model updating. The y-axis gives percent confidence, as described in the text. The red vertical line denotes the point at which an 
update is triggered by our model uncertainty quantification mechanism.}
\label{fig:UQ-BraggNN}
 \end{figure}

We show in \figautoref{fig:UQ-BraggNN}, as the black line labeled ``Before Trigger,'' certainty calculated in this way when the embedding and clustering models trained on the first five datasets are used for all subsequent datasets.
We also show, as the red line labeled ``After Trigger,'' the model certainty observed when \proj{} is configured to trigger the control plane functionality, and thus retrain the embedding module, re-train the clustering algorithm, and update the data store, when certainty drops below 80\%. 
We see that the embedding and clustering models trained on data from the first five experiments perform consistently well (the black line) until dataset 23, when we observe a significant drop in the amount of data being assigned to their respective clusters, from 97\% to below 60\%. 
The retrained model, on the other hand,  consistently assigns the data from different stages of experiments to their respective clusters with high certainty (the red line)---demonstrating the ability of \proj{} to adapt to dynamic experimental conditions and dataset characteristics.

%% file: discussion.tex
\section{Discussion}
\label{limits}
\reviews{
We discuss how \proj{} can be employed to achieve the goal of real-time performance at high-data-rate instruments.
}

\reviews{\textbf{Application:}
We have reported here on results that demonstrate \proj{}'s applicability to varying data distributions in several different experimental set ups. Importantly, \proj{} assumes that there will be a correlation between the experimental data collected in different experimental settings. If data generated and gathered during various experimental settings do not demonstrate such correlations, the use of \proj{} may result in sub-par or no improvement in end-to-end ML model training time.}
 
\reviews{
\textbf{Generalizability:} We used three computer-vision-related applications from HEDM (at APS and LCLS) to evaluate \proj{} performance. Similarly, we evaluated the performance in an HPC environment (Argonne Leadership Computing Facility). In future work we plan to explore a broader set of machine learning applications that includes natural language processing and non scientific applications. We also plan to evaluate performance in a cloud environment and to further study the scalability of \proj{}.
}

\reviews{
\textbf{An example of failure:} 
We initially used an autoencoder to build embedding for Bragg peaks, as we had previously used successfully for CookieBox data. 
However, this approach did not work well for indexing trained models, primarily because the autoencoder, being tasked to reconstruct the image from latent variables, proved to be overly sensitive to pixel-wise differences. 
BraggNN is concerned only with finding the center of mass in a supplied image, and two peaks can be considered as identical from the physics point of view if one is just a rotation (i.e., augmentation) of the other---a common situation in diffraction experiments. 
We thus came up with the BYOL~\cite{grill2020bootstrap} method which can be trained to be agnostic to augmentations inspired by physics (e.g., rotation in-variance) and experimental conditions (e.g., prior of the noise model). 
}



%% file: related.tex

\section{Related Work}\label{related}

Our work in \proj{} relates to the fields of representation learning, storage of ML training data, transfer learning, and trained model indexing and recommendation. We highlight here the state-of-the-art in each field and compare and contrast with \proj{}.

Self-supervised learning is a powerful method for learning useful representations without supervision from labels that can greatly reduce the performance gap between supervised models on various downstream vision tasks. Self-supervised representation learning methods are being widely adopted due to their effectiveness in a wide variety of applications, such as relative patch prediction \cite{doersch2015unsupervised}, solving jigsaw puzzles \cite{noroozi1603unsupervised}, colorization \cite{zhang2016colorful}, and rotation prediction \cite{gidaris2018unsupervised, chen2019self,chen2020simple}. 

Some early work in unsupervised learning from images aimed to discover object categories by using hand-crafted features and various forms of clustering (e.g., for learning a generative model over bags of visual words~\cite{russell2006using,sivic2005discovering}). More recent and widely used autonomic algorithms for feature representations include denoising autoencoders \cite{bengio2014deep,wang2015unsupervised}, sparse autoencoders \cite{le2013building}, and skip-gram models \cite{mikolov2013distributed}. Many related methods rely on contrastive learning (e.g., SimCLR \cite{chen2020simple}, MoCO \cite{chen2021empirical}), which learns representations by maximizing agreement between differently augmented views of the same data example. In particular, this technique contrasts positive pairs against negative pairs and minimizes differences between positive pairs to avoid collapsing solutions \cite{tian2021understanding,ye2019unsupervised}. Another recent line of work (e.g., BYOL~\cite{grill2020bootstrap}, SimSiam \cite{chen2021exploring}) employs asymmetry of the learning update (stop-gradient operation) to avoid trivial solutions. We employ these methods in \proj{} for efficient pseudo-labelling of input data. 

State-of-the-art storage systems (e.g., Diesel~\cite{wang2020diesel}, TableFS~\cite{ren2012tablefs}, DeltaFS~\cite{zheng2018scaling}, IndexFS~\cite{ren2014indexfs}) make extensive use of metadata to improve throughput. 
These systems employ caching to improve the throughput of distributed file system during the training process using metadata information. This research has focused primarily on reducing I/O bottlenecks during the distributed ML model training phase. The goal of \proj{} is different. To manage ever-growing training data that are generated at ultra-high speeds from scientific experiments, we need a specialized system for data management that provides fast lookup by using a specialised indexing mechanism.  

From the ML systems prospective, \proj{} is closely related to continuous training systems in which models are trained continuously to deal with dynamically evolving datasets generated during scientific experiments. 
Liu et al.~\cite{liu2021bridge} proposed an end-to-end ML system for scientific applications. In serial crystallography~\cite{Nass2020,Meents2017}, Bragg spots from past experiments are used to train detectors continuously for use in the current experiment. 
PtychoNN~\cite{cherukara2020ai} is proposed for  phase retrieval in ptychography, using data generated in the early stages of experiments. 
In all of these frameworks, models are repeatedly trained from scratch. 
In contrast, \proj{} leverages historical data through pseudo-labeling to reduce data annotation costs. 
The \mms{} enables model updating by fine-tuning of historical models identified by indexing their training dataset.